\documentclass[letterpaper]{article} 
\usepackage{aaai2027}  
\usepackage[hyphens]{url}  
\usepackage{graphicx} 
\usepackage{natbib}  
\usepackage{caption} 
\usepackage{algorithm}
\usepackage{algorithmic}
\usepackage{amssymb}

\usepackage{newfloat}
\usepackage[table]{xcolor}
\usepackage{amsmath}
\usepackage{listings}
\DeclareCaptionStyle{ruled}{labelfont=normalfont,labelsep=colon,strut=off} 
\floatstyle{ruled}
\newfloat{listing}{tb}{lst}{}
\floatname{listing}{Listing}

\usepackage{booktabs}
\usepackage{colortbl}

\title{Driving on Registers, Reasoning on Risk: Risk-Aware Occupancy for Register-Based End-to-End Autonomous Driving}

\author{
Jiaxing Chen\textsuperscript{1},
Hengduo Zou\textsuperscript{1},
Yukai Qin\textsuperscript{1},
Yiren Zhao\textsuperscript{2},
Lidong Yu\textsuperscript{3}
Bolin Gao\textsuperscript{1}\thanks{$\dagger$ Corresponding author: Bolin Gao}
}
\affiliations{
\textsuperscript{1}School of Vehicle and Mobility, Tsinghua University\\
\textsuperscript{2}The Hong Kong University of Science and Technology (Guangzhou)\\
\textsuperscript{3}Neolix, Beijing, China}

\nocopyright

\begin{document}

\maketitle

\begin{abstract}
Multimodal trajectory prediction improves behavioral coverage in end-to-end autonomous driving, but existing methods remain limited by sparse scene representations. Incomplete evidence leads to low-quality candidate generation and unreliable ranking among geometrically similar trajectories. On a register-based baseline, bad and poor candidates constitute 19.74\% of the candidate set, while the oracle-best candidate ranks only 33.9th on average. We propose RRDrive, which introduces risk-aware occupancy as a dense, temporally aligned, and trajectory-queryable representation. Its global structure guides high-quality multimodal generation, while candidate-conditioned risk queries support fine-grained selection. We further construct RiskOcc-NAVSIM with automatic risk annotations. RRDrive achieves a selected-trajectory PDMS of \textbf{0.951}, representing a 1.5\% relative improvement over the baseline (0.937), and improves the average candidate PDMS by 7.7\%. In challenging scenes, it improves candidate PDMS by 30.2\% and increases the Spearman correlation among good candidates by 0.41, from 0.26 to 0.67. To move beyond this oracle setting, we further develop an external RiskOcc predictor, a perception module that estimates risk-aware occupancy directly from sensor inputs. The competitive performance validates the representation’s feasibility.

\end{abstract}

\section{Introduction}

End-to-end autonomous driving must produce a single executable trajectory from uncertain observations. Multimodal trajectory prediction improves robustness by preserving multiple plausible driving behaviors before final selection. However, existing methods still suffer from low-quality candidate generation and unreliable trajectory ranking. We argue that these failures are not inherent to multimodality, but largely stem from sparse scene representations. Agent- and map-based features provide compact structural information, yet omit continuous evidence about unstructured obstacles, road boundaries, uncertain regions, and future interactions. Consequently, they provide insufficient global guidance for candidate generation and inadequate local risk evidence for a risk-aware selector.

\begin{figure}[t!]
\centering
\includegraphics[width=0.47\textwidth,       
        trim=1 1 1 1,
        clip]{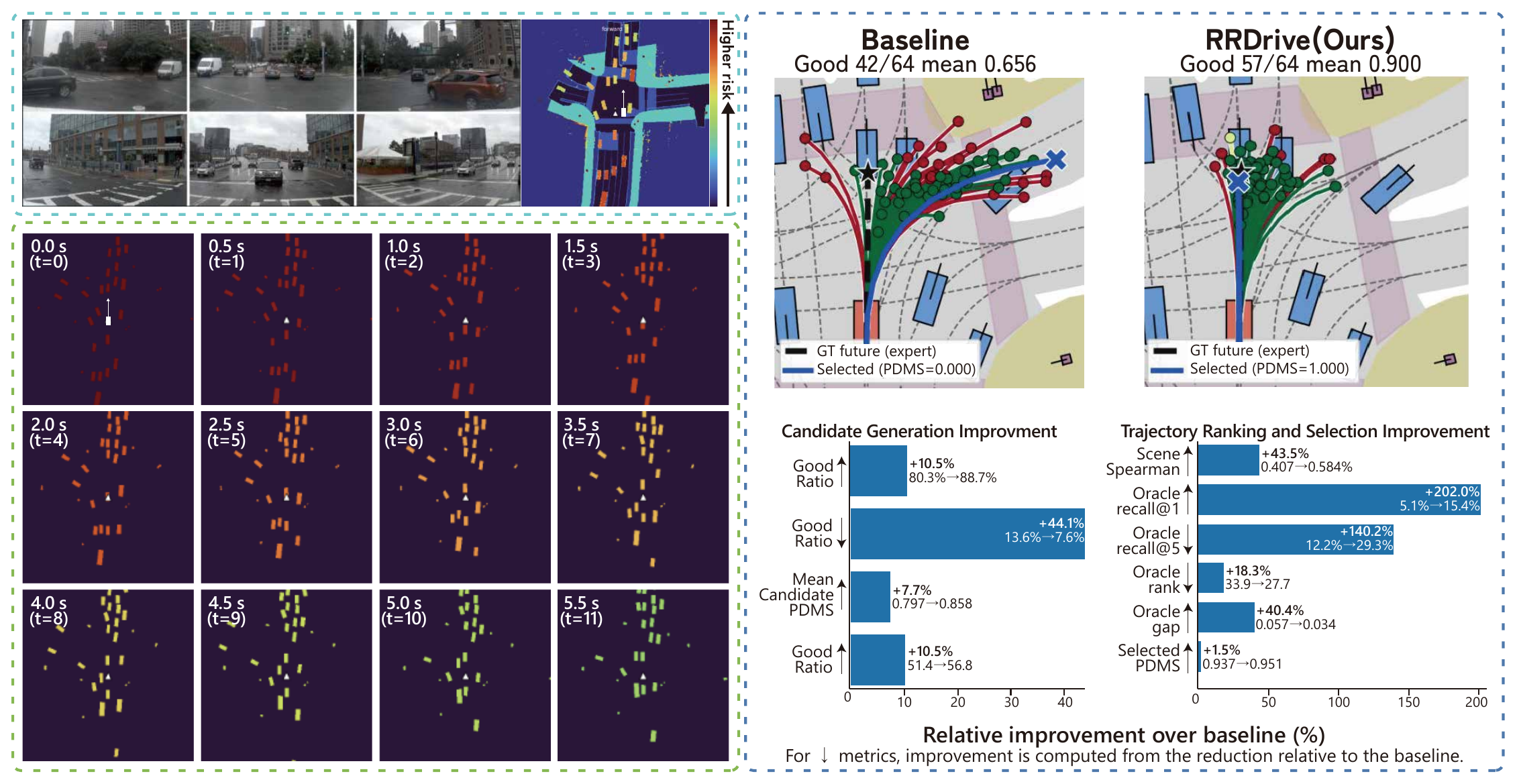} 
\caption{Motivation and overall performance of RRDrive.
    Register-based baselines face both candidate generation waste and ranking errors in complex scenes, driven by unmodeled local future risks.
    Left: Multi-view inputs and temporal evolution of risk-aware occupancy.
    Right: Trajectory comparison and quantitative improvements validate our dense trajectory-queryable risk representation.}
\label{fig:teaser}
\end{figure}

We analyze this limitation on baseline, a register‑based planner with separate trajectory generation and scoring \cite{drivor}.
As shown in Figure~\ref{fig:teaser}, a single complex driving scene exhibits two coupled failure modes.
The generator outputs abundant unsafe, off‑route, or ineffective candidates that exhaust the limited candidate budget.
Meanwhile, geometrically similar good trajectories form dense bundles; their ground‑truth oracle PDMS differs from localized future risks, leading to mis‑ranking by the predicted scorer.
A missing good candidate cannot be recovered by the scorer, while incorrect ranking discards already‑available high‑quality solutions.
These observations suggest that multimodal generator‑selector architectures require a dense representation supporting both global risk reasoning and trajectory‑specific evaluation.

The risk-aware occupancy, an explicit, dense, and trajectory-queryable representation in the ego-centric BEV form, proposed by \cite{riskoccupancy}. It unifies map-derived static risks and dynamic occupancy risks within a common field. Its global structure exposes feasible and hazardous regions, enabling the generator to allocate candidates toward high-quality behaviors. Its local values can be queried along each trajectory to form a candidate-specific risk profile, revealing differences in collision exposure, road-boundary risk, and future interactions. Risk-aware occupancy therefore provides the complementary information required for both candidate generation and fine-grained selection.

\begin{figure*}[t]
    \centering
    \includegraphics[width=0.99\linewidth
    ]{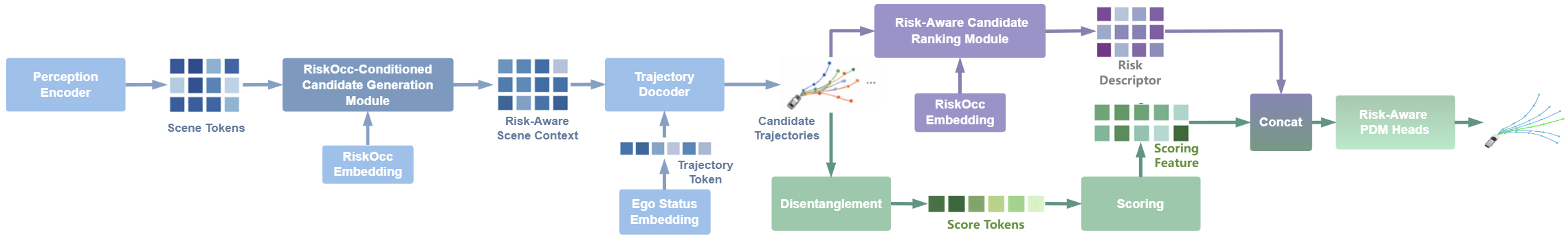}
    \caption{Overall architecture of RRDrive. A risk tokenizer encodes BEV risk-aware occupancy, which is fused with camera scene tokens for risk-conditioned candidate generation and read out along each candidate trajectory for risk-aware selection.}
    \label{fig:overall}
\end{figure*}

Building upon this representation, we present a Risk‑Register E2E framework driven by global‑local risk‑aware occupancy cues (RRDrive) (Figure~\ref{fig:overall}), a multimodal end‑to‑end planner featuring global risk fusion and trajectory‑conditioned risk extraction. Global risk fusion incorporates the complete risk field into scene features to improve candidate generation. Trajectory-conditioned extraction samples local risks along each proposal and augments the scorer with explicit risk evidence. We further introduce a risk-aware occupancy prediction branch to estimate the representation from sensor inputs. Because existing datasets do not provide the required supervision, we develop an automatic annotation pipeline and construct RiskOcc-NAVSIM. We also adopt targeted metrics to separately evaluate candidate quality and ranking reliability.

Experiments on NAVSIM show that RRDrive improves both capabilities. It raises the overall average candidate PDMS from 0.797 to 0.858 and the selected-trajectory PDMS from 0.9367 to 0.9506. In generation-challenging scenes, the average candidate PDMS increases from 0.5395 to 0.7026, while the proportion of good candidates rises from 50.13\% to 69.76\%. In selection-challenging scenes, the Spearman correlation among good candidates improves from 0.2646 to 0.6745, and oracle-best recall increases from 3.62\% to 24.86\%. These results demonstrate that explicit risk representation improves both multimodal trajectory quality and best-trajectory selection.

The main contributions are summarized as follows:

\begin{itemize}
    \item We identify compact scene register representation as the shared bottleneck of multimodal trajectory generation and selection, and exploit risk‑aware occupancy for global‑local collaborative reasoning: global risk guides feasible candidate allocation, while trajectory‑conditioned local risk queries offer discriminative cues for ranking geometrically similar trajectories.
    \item We propose RRDrive, a register‑based end‑to‑end driving architecture that integrates risk‑conditioned candidate generation and trajectory‑conditioned risk‑aware selection. It substantially improves candidate‑pool quality and ranking fidelity, yielding consistent gains in challenging interactive scenarios.
    \item We build RiskOcc4D‑NAVSIM by automated annotation of NAVSIM to furnish ground‑truth supervision and an evaluation testbed for risk‑aware occupancy. We further develop an external RiskOcc predictor producing lossy risk maps from sensor inputs to validate pipeline robustness under non‑oracle perceptual conditions.
\end{itemize}

\section{Related Work}

Existing end-to-end autonomous driving planners fall into
three paradigms by trajectory candidate construction and selection: deterministic point estimation, discrete hypothesis
scoring, and continuous distribution modeling. Each has clear
performance trade-offs and inherent flaws. We take the practical discrete candidate scoring paradigm as baseline, targeting
its core defect: implicit modeling of scene-trajectory matching. Represented by UniAD\cite{hu2023planningorientedautonomousdriving} and TransFuser\cite{chitta:22transfuser},
deterministic single-trajectory regression adopts end-to-end
networks with behavior cloning to output one trajectory per
forward pass. Simple, low-latency and multi-task trainable, it
fits real-time onboard deployment. Yet it frames planning as
point estimation and cannot preserve multiple multimodal
feasible trajectories. In interactive scenarios with diverse
plausible maneuvers, it outputs averaged, conservative suboptimal trajectories with weak decision diversity and poor
long-tail robustness.

Discrete candidate generation with learnable scoring
(VADv2\cite{jiang2026vadv2endtoendvectorizedautonomous}, SparseDrive\cite{sun:24sparsedrive}) produces discrete trajectory
candidates via sparse queries and structured anchors, then
ranks trajectories via a dedicated scoring network. It resolves the diversity deficit of single-trajectory regression and
avoids generative models’ iterative sampling cost, balancing multi-hypothesis reasoning and fast inference. Its key
bottleneck lies in globally compressed implicit scene features for trajectory evaluation: candidate-environment and
interaction boundary matching is only implicitly inferred
without explicit modeling or precise querying. This yields
abundant low-quality redundant samples and unstable fine-grained ranking among similar high-quality trajectory candidates, capping performance in complex interactive scenarios.

Generative sampling with Best-of-N selection (DiffusionDrive \cite{liao2025diffusiondrivetruncateddiffusionmodel}, GoalFlow \cite{xing2025goalflowgoaldrivenflowmatching}, TrajFlow \cite{li2026trajflownationwidepseudogps}) learns scene-conditioned continuous trajectory distributions to break discrete candidates’
expressiveness limits, generating highly diverse, scene-wide
trajectory samples better suited to long-tail complex interactions. DiffusionDrive leverages anchor priors and truncated
diffusion sampling to cut sampling costs while enabling multimodal trajectories. Still, iterative sampling brings heavy
inference overhead incompatible with onboard real-time demands; random sampling lacks stability, and trajectory quality heavily relies on post-hoc scoring. Without explicit spatiotemporal environmental constraints, Best-of-N filtering
fails to accurately distinguish near-identical high-quality candidates, hurting deployment stability.

\section{Method}

Figure~\ref{fig:overall} gives an overview of RRDrive. On top of a register-based generator--selector planner, risk-aware occupancy enters through two pathways: a risk-conditioned generation module that fuses the global risk field into scene features to steer candidate generation, and a risk-aware selection module that reads the field along each candidate to sharpen ranking. A label-construction pipeline supplies theF ground-truth risk representation for training, and an external predictor estimates it from sensors for sensor-only inference. The four subsections below detail these components in turn.

\subsection{Risk-Conditioned Candidate Generation}
\begin{figure}[ht]
    \centering
    \includegraphics[width=0.99\linewidth,
            ]{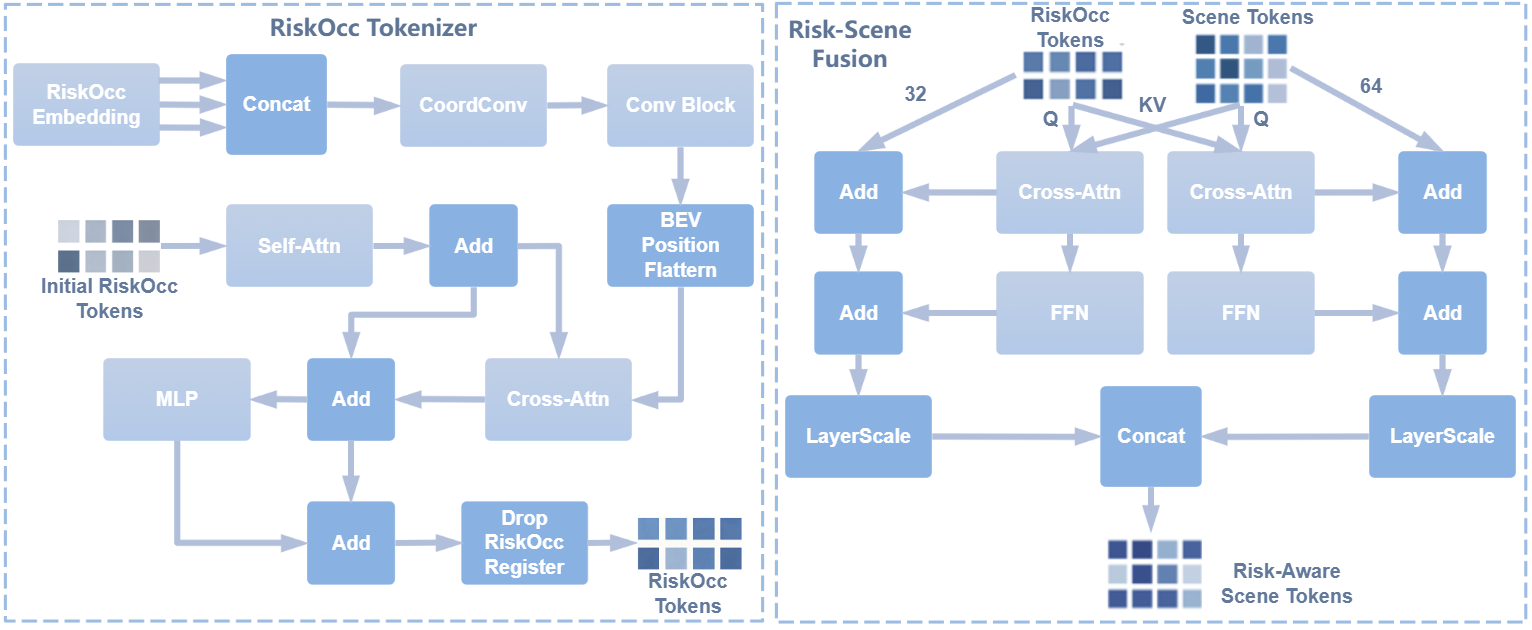}
    \caption{RiskOcc-conditioned candidate generation module. Risk tokens and camera scene tokens are combined by bidirectional cross-attention with LayerScale gating, and the fused context conditions the trajectory decoder so that the finite candidate budget is steered toward feasible, high-quality behaviors.}
    \label{fig:RiskOcc-Conditioned}
\end{figure}

A finite budget spent on unsafe or off-route proposals sets a quality floor that no downstream scorer can recover, so this module makes scene-level risk explicit and conditions candidate generation on it. The framework represents planning as scoring a finite candidate set $\mathcal{T}=\{\tau_i\!\}_{i=1}^{N}$, where each proposal $\tau_i=(x_{i,t},\!y_{i,t},\!\theta_{i,t})_{t=1}^{T_p}$ is an ego-centric future trajectory. In the original design, candidates come from visual scene tokens and trajectory features alone; because this implicit representation cannot separate feasible regions from hazardous ones, the budget leaks toward such proposals, which the explicit RiskOcc field is designed to prevent. Figure~\ref{fig:RiskOcc-Conditioned} illustrates this module.

We use BEV risk-aware occupancy $\mathcal{R}=\{R^s,\!\{R^d_t\!\}_{t=1}^{T_r},\!R^l\}$, where $R^s\in\mathbb{R}^{H\!\times\!W}$ is global scene RiskOcc, $R^d_t\in\mathbb{R}^{C_d\!\times\!H\!\times\!W}$ is time-indexed dynamic risk with velocity cues, and $R^l\in\mathbb{R}^{H\!\times\!W}$ is lane-boundary risk, all in the trajectory frame. Rather than a handcrafted cost map, $\mathcal{R}$ is encoded into a compact token set by a risk tokenizer $\mathcal{E}_\phi$,
\begin{equation}
Z_r=\mathcal{E}_\phi\!\big(\textnormal{Concat}(R^s,\!\{R^d_t\!\}_t,\!R^l)\big)\in\mathbb{R}^{K\!\times\!D},
\end{equation}
built from a convolutional BEV stem with coordinate-aware priors, residual channel reweighting, and Perceiver-style cross-attention onto $K$ learnable queries, preserving safety-critical local cues.

Given camera scene tokens $Z_c$, a risk-aware context is formed by bidirectional cross-attention $\mathrm{CA}(\cdot,\cdot)$,
\begin{equation}
\begin{aligned}
\tilde{Z}_c &= Z_c + \gamma_{c\leftarrow r}\,\mathrm{CA}(Z_c,Z_r),\\
\tilde{Z}_r &= Z_r + \gamma_{r\leftarrow c}\,\mathrm{CA}(Z_r,Z_c),
\end{aligned}
\end{equation}
whose LayerScale coefficients $\gamma$ are initialized near zero, so fusion is identity-preserving at initialization and warm-starts stably. The fused context $M=[\tilde{Z}_c,\tilde{Z}_r]$ is cross-attended by the trajectory decoder to produce $\mathcal{T}$, letting global risk structure steer the budget toward feasible behaviors while suppressing unsafe or off-route proposals. The same $M$ is shared with the selection scorer described next, so risk shapes both candidate representation and final ranking.

\subsection{Risk-Aware Best-Candidate Selection}
\label{subsec:selection}
\begin{figure}[ht]
    \centering
    \includegraphics[width=0.99\linewidth
    ]{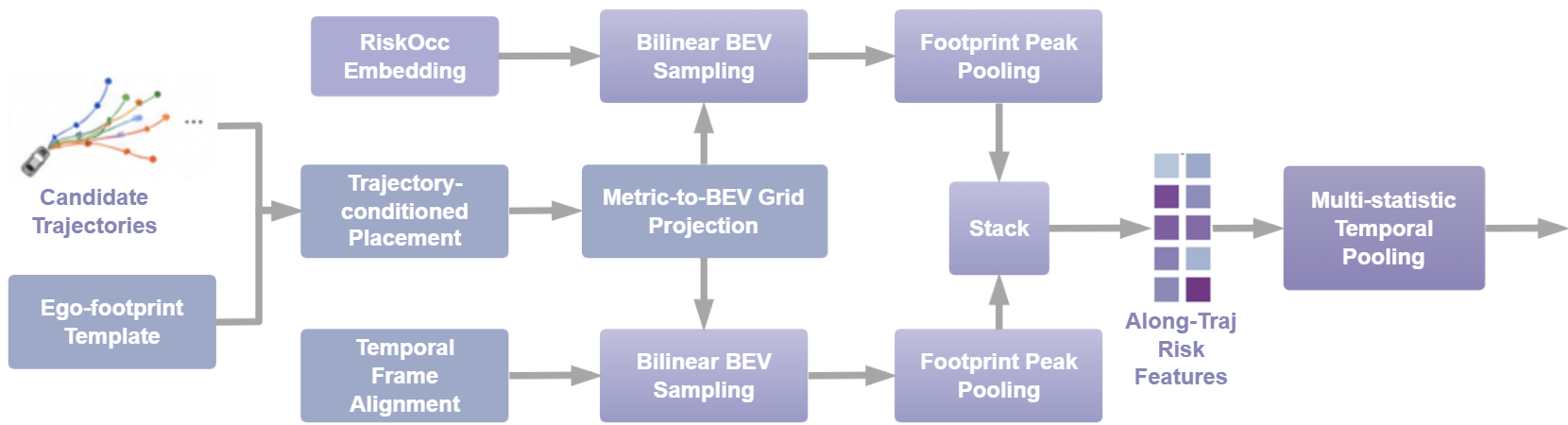}
    \caption{Risk-aware candidate ranking module. For each candidate, the ego footprint samples the time-aligned risk field along the trajectory to form a local risk profile, which augments the risk-sensitive scorer heads and sharpens ranking among geometrically similar good candidates.}
    \label{fig:Risk-Aware}
\end{figure}

Even a strong scorer mis-ranks geometrically similar good candidates when risk stays implicit, so this module hands the scorer a trajectory-conditioned readout of the RiskOcc field to expose the fine progress--risk margin that separates them. For each candidate, scorer attention reads the shared context, $u_i=\mathrm{Attn}(q_i,M)$. Because global tokens alone cannot reveal which risk regions a proposal traverses, we add a trajectory-conditioned local query. At waypoint $t$, the ego footprint $\mathcal{B}(\tau_{i,t})$ is placed on the BEV grid and bilinearly samples the time-aligned map $\bar{R}_t=[R^s,R^d_{\alpha(t)},R^l]$, where $\alpha(t)$ aligns waypoint $t$ to dynamic frame $t$ (clamped to $T_r{-}1$). For channel $c$, exposure is a top-$k$ mean over footprint points,
\begin{equation}
\begin{aligned}
\mathcal{P}_{i,t,c}&=\mathrm{TopK}_{k}\{\bar{R}_{t,c}(q):q\in\mathcal{B}(\tau_{i,t})\},\\
\rho_{i,t,c}&=\tfrac{1}{k}\sum\nolimits_{p\in\mathcal{P}_{i,t,c}}\bar{R}_{t,c}(p),
\end{aligned}
\end{equation}
which retains local peaks that plain averaging would dilute. The per-waypoint sequence is summarized into a fixed descriptor
\begin{equation}
\begin{aligned}
g_i=\mathrm{Concat}\big(&\operatorname{mean}_t\rho_{i,t},\ \max_t\rho_{i,t},\ \min_t\rho_{i,t},\ \rho_{i,T_p}\big),
\end{aligned}
\end{equation}
capturing average exposure, closest-approach risk, low-risk feasibility, and terminal risk. For each submetric $m\in\{\mathrm{NC},\mathrm{DAC},\mathrm{DDC},\mathrm{TTC},\mathrm{EP},\mathrm{C}\}$, the descriptor augments only the risk-aware heads $\mathcal{M}_{\mathrm{risk}}$, while the rest use $u_i$ alone,
\begin{equation}
h_i^m=
\begin{cases}
f_m([u_i,g_i]), & m\in\mathcal{M}_{\mathrm{risk}},\\
f_m(u_i), & \text{otherwise}.
\end{cases}
\end{equation}
Let $\mathcal{M}_{\mathrm{p}}=\{\textnormal{NC},\!\textnormal{DAC},\!\textnormal{DDC}\},\quad \mathcal{M}_{\mathrm{w}}=\{\textnormal{TTC},\!\textnormal{EP},\!\textnormal{C}\}$. Selection keeps the PDM-style weighted-log aggregation,
\begin{equation}
\begin{aligned}
s_i&=\sum_{m\in\mathcal{M}_{\mathrm{p}}} w_m\log p_i^m
+\log\!\Big(\sum_{m\in\mathcal{M}_{\mathrm{w}}} w_m\,p_i^m\Big),\\
i^\star&=\arg\max\nolimits_i s_i.
\end{aligned}
\end{equation}
This preserves the baseline's multi-submetric structure and avoids a manual risk penalty. Since simple statistics such as risk mean or max correlate only weakly with true PDMS and safety submetrics, RiskOcc acts not as a linear surrogate but as trajectory-aligned nonlinear evidence for the neural scorer. Figure~\ref{fig:Risk-Aware} depicts this ranking module.

For the refinement phase, the proposal generator and trajectory decoder are frozen, and only the risk encoder, scorer attention, and the TTC/EP heads are updated. The objective keeps the submetric BCE terms and adds continuous smooth-$\ell_1$ calibration ($\mathrm{SL}_1$) for the two risk-sensitive submetrics,
\begin{equation}
\begin{aligned}
\mathcal{L} &= \sum_m \lambda_m\mathcal{L}_{\mathrm{BCE}}(p^m,y^m) + \lambda_{\mathrm{TTC}}\,\mathrm{SL}_1(p^{\mathrm{TTC}},y^{\mathrm{TTC}}_{\mathrm{c}}) \\
&\quad + \lambda_{\mathrm{EP}}\,\mathrm{SL}_1(p^{\mathrm{EP}},y^{\mathrm{EP}}_{\mathrm{c}}).
\end{aligned}
\end{equation}
Although the non-target heads are frozen, their BCE losses still backpropagate through the shared scorer features, anchoring NC, DAC, DDC, and comfort during refinement. The model thus does not relearn the planner but sharpens its interpretation of the RiskOcc--trajectory relation, concentrating gains in fine-grained ranking among high-quality candidates—particularly TTC-sensitive exposure and progress-aware selection once safety is satisfied—rather than inducing conservative behavior.

\subsection{Risk-Aware Occupancy Generation}
\label{sec:label}

Both pathways above learn from a ground-truth RiskOcc field, so this module constructs that supervision itself as the RiskOcc-NAVSIM labels. We distill dense risk-aware occupancy representation from NAVSIM on a shared
ego-centric BEV lattice of $H\times W=400\times400$ grid cells at resolution
$\Delta=0.25\,\mathrm{m}$ (covering $[-50,50)^2\,\mathrm{m}$), with risk normalized
to $r\in[0,1]$. An ego-frame point $(x,y)$ maps to the grid cell
$(i,j)=(\lfloor(y{-}y_{\min})/\Delta\rfloor,\ \lfloor(x{-}x_{\min})/\Delta\rfloor)$
with $x_{\min}{=}y_{\min}{=}{-}50$ and metric grid cell center $p_{ij}$. Three pipelines
yield complementary targets: a map-derived \emph{lane-marking RiskOcc map},
a point-cloud \emph{global RiskOcc map} refined by that prior,
and an annotation-derived \emph{dynamic RiskOcc map}; the last
two are stacked into the final label.

\subsubsection{Map-Derived Risk-Aware Occupancy}
\label{sec:lane}
From the vectorized HD map, we rasterize a single-channel prior
$M\in\mathbb{R}^{H\times W}$, where each grid cell inherits the value of the top-most
(highest-priority) layer covering it:

\begin{equation}
M(i,j)=\phi(\ell^\star),\quad
\ell^\star=\arg\max_{\ell\in\mathcal{L}}\pi(\ell)\,\mathbb{1}[p_{ij}\in\ell],
\end{equation}

with render priority $\pi(\cdot)$ and coverage indicator $\mathbb{1}[\cdot]$. The
layer values $\phi$ are drivable interior $0$, background $0.05$,
crosswalk/stop-line/lane boundary $0.1$, drivable boundary $0.2$, walkway $0.3$,
and lane-connector centerline $-0.1$. $M$ is reused in the global map pipeline.

\subsubsection{Global Risk-Aware Occupancy}
\label{sec:global}
The semantic point cloud is voxelized into a $400\times400\times16$ grid
($z\in[-2,2)$); a voxel is occupied if it holds more than two points, taking the
majority semantic label $V$, else background. For each BEV grid cell, we scan its
column upward from the ground ($z{=}0$) for the first salient voxel $w^\star$
(neither road nor background) at height index $\eta$ (larger $\eta$ = taller).
With normalized height $\hat h=\Delta\eta/2.5$ and
$g=\operatorname{clip}(0.8\hat h+0.2,0,1)$, the point-cloud risk is
$R_{\mathrm{pc}}=1$ if $V(w^\star)=2$, $R_{\mathrm{pc}}=g$ if $V(w^\star)\in\{1,3\}$,
and $0$ otherwise. We fuse with the lane prior by element-wise maximum ($\vee$) and
refine with one grayscale morphological closing ($3\times3$ element $B$),
$R=((R_{\mathrm{pc}}\vee M)\oplus B)\ominus B$. Finally, $R$ is quantized into $11$
ordinal classes: class $0$ for $R<0.05$, class $10$ for free background
($R{=}0.05$), and classes $1$--$9$ for monotonically increasing risk over
$(0.05,1]$.

\subsubsection{Dynamic Risk-Aware Occupancy}
\label{sec:dynamic}
We produce a spatio-temporal tensor of $T=12$ frames (index $k\in\{0,\dots,11\}$ at
$\delta t=0.5\,\mathrm{s}$, i.e.\ $t=k\,\delta t\in\{0,\dots,5.5\}\,\mathrm{s}$), all
expressed in the $t{=}0$ ego frame. Only agents of $\{$vehicles, pedestrians, and
bicycles$\}$ visible $t{=}0$ are kept, with short gaps ($\le2$ consecutive
missing frames) linearly interpolated. Each agent carries a velocity clamped and
normalized into the $t{=}0$ frame,
$\tilde v=\operatorname{clip}(v/20,-1,1)$, $s=\lVert\tilde v\rVert_2$, and a
frame-indexed risk that decays over the horizon and grows with speed:
\begin{equation}
\rho_a^{\,k}=\min\!\Big((1-0.05\,k)+0.05\,s/\sqrt2,\ 1\Big).
\end{equation}
An agent with center $c_a^k$, size $(\ell_a,w_a)$ and heading unit vector
$\mathbf{u}_\theta$ occupies the grid cell $(i,j)$ when the grid cell center $p_{ij}$ lies inside
its oriented box.
Overlaps are resolved by priority, keeping the higher-risk agent and breaking
ties by the smaller instance identity,
$a^\star=\arg\max_{a:\,B_a^{\,k}(i,j)=1}(\rho_a^{\,k},\,-\mathrm{id}_a)$---so each
occupied grid cell stores the winning agent's risk $\rho_{a^\star}^{\,k}$ and
velocity $(\tilde v_x,\tilde v_y)$, while free grid cells are zero.

\subsection{External RiskOcc Predictor}

\subsubsection{Formulation}
Injecting a temporal, BEV-based occupancy branch inside our planner would break
its lightweight, history-free, PV register-token design, so we keep risk
perception external as a standalone predictor that serves only as a probe of the
planner's robustness to lossy predicted RiskOcc. Given a short sensor observation
$\mathcal{O}$, this predictor $f_\theta$ emits a BEV RiskOcc field of a static
map, a lane map, and a dynamic general-movable-object (GMO) sequence with per-cell motion,
\begin{equation}
\begin{aligned}
f_\theta(\mathcal{O}) = \big(\,
  &\mathbf{R}^{s},\,\mathbf{R}^{\ell}\in[0,1]^{H\times W},\\[-1pt]
  &\mathbf{R}^{d}\in[0,1]^{T\times H\times W},\;
   \mathbf{V}\in\mathbb{R}^{T\times H\times W\times 2}\,\big),
\end{aligned}
\end{equation}
with $H=W=400$, $T=12$, through a two-stage encoder--decoder
$\mathcal{O}\xrightarrow{\Phi}\mathbf{B}\xrightarrow{\Psi}(\cdot)$ with fused BEV
$\mathbf{B}\in\mathbb{R}^{256\times 50\times 50}$, and $\odot$ denotes the Hadamard product.

\subsubsection{Architecture}
The encoder $\Phi$ outputs a BEV feature $\mathbf{B}$ by fusing two streams: an
ego-aligned motion history summarized by a selective state-space operator, and a
multi-height deformable camera-to-BEV lift following FusionAD~\cite{fusionad}. The
streams are merged by residual spatial cross-attention, with modality dropout for
robustness to missing sensors. The decoder $\Psi$ upsamples $\mathbf{B}$ with skip
fusion and predicts static and lane risk through class-first heads (a class map
that conditions a risk regressor).

The dynamic branch operates on a stop-gradient BEV context $\mathbf{P}$. For each
future frame $t$, a per-frame query $\mathbf{q}_t$ drives a deformable forecaster
$\mathbf{F}_t=\mathcal{D}(\mathbf{q}_t,\mathbf{P})$; a FiLM module conditioned on an
occupancy prior (AdaNorm) refines it into $\tilde{\mathbf{F}}_t$ through a
LayerScale residual, and a dense $K\!=\!12$ head maps $\tilde{\mathbf{F}}_t$ to
per-level logits $\mathbf{z}^{d}_t$. The exported dynamic risk is the softmax-class
expectation over the discrete levels $\{r_c\}_{c=1}^{K}$,
\begin{equation}
\mathbf{R}^{d}_t=\sum\nolimits_{c=1}^{K}\operatorname{softmax}(\mathbf{z}^{d}_t)_c\,r_c,
\end{equation}
and the velocity field is gated by the same expectation,
$\mathbf{V}_t=\mathbf{R}^{d}_t\odot\operatorname{Vel}(\tilde{\mathbf{F}}_t)$ with
$\operatorname{Vel}$ a linear readout. This continuous expectation, rather than a
hard threshold, preserves near-frame accuracy, while a clean-label ground truth
retaining only $t\!=\!0$-visible targets suppresses spurious far-frame occupancy.

\section{Experiments}

RRDrive closes the two bottlenecks raised in the abstract---the low-quality floor of the candidate set and the unreliable fine-grained ranking of near-identical good candidates---most decisively where the baseline is weakest. Each claim has a dedicated experiment: candidate generation (Table~\ref{tab:gen}), scene-level ranking (Tables~\ref{tab:rank} and~\ref{tab:probe}), the concentration of both gains in the hardest scenes (Table~\ref{tab:difficulty}), per-component attribution (Table~\ref{tab:ablation}), and benchmark standing (Table~\ref{tab:navsim_compare_sorted}).

\begin{table}[h!]
\centering
\small
\setlength{\tabcolsep}{1pt}
\begin{tabular}{l l c  c c c c}
\toprule
\textbf{Methods}  & \textbf{PDMS$\uparrow$} & \textbf{NC$\uparrow$} & \textbf{DAC$\uparrow$} & \textbf{EP$\uparrow$} & \textbf{TTC$\uparrow$} & \textbf{Comf.$\uparrow$} \\
\midrule
\multicolumn{7}{l}{\textbf{Baselines \& Oracle}} \\
\rowcolor{gray!20} Human Driver \citeyear{dauner2024navsimdatadrivennonreactiveautonomous}  & 94.8 & 100.0 & 100.0 & 87.5 & 100.0 & 99.9 \\
PDMS-Closed \citeyear{dauner2023partingmisconceptionslearningbasedvehicle} & 89.1 & 94.6 & 99.8 & 99.9 & 86.9 & 89.9 \\
Ego-stat. MLP \citeyear{dauner2024navsimdatadrivennonreactiveautonomous} & 65.6 & 93.0 & 77.3 & 62.8 & 83.6 & \textbf{100.0} \\
\midrule
\multicolumn{7}{l}{\textbf{VLA-based Methods}} \\
UniVLA \citeyear{bu2025univlalearningacttaskcentric} &  81.7 & 96.9 & 91.1 & 76.8 & 91.7 & 96.7 \\
FSDrive \citeyear{zeng2025futuresightdrivethinkingvisuallyspatiotemporal} & 85.1 & 98.2 & 93.8 & 80.1 & 93.3 & 99.9 \\
AutoVLA \citeyear{zhou2025autovlavisionlanguageactionmodelendtoend} & 89.1 & 98.4 & 95.6 & 81.9 & \textbf{98.0} & 99.9 \\
DriveVLA-W0 \citeyear{li2025drivevlaw0worldmodelsamplify}  & 90.2 & 98.7 & \textbf{99.1} & 83.3 & 95.3 & 99.3 \\
SpanVLA \citeyear{zhou2026spanvlaefficientactionbridging} & 90.3 & 99.1 & 97.1 & 86.3 & 95.2 & \textbf{100.0} \\
ReCogDrive \citeyear{li2025recogdrivereinforcedcognitiveframework} & 90.8 & 97.9 & 97.3 & 87.3 & 94.9 & \textbf{100.0} \\
SGDrive \citeyear{li2026sgdrivescenetogoalhierarchicalworld} & 91.1 & 98.6 & 97.8 & 85.8 & 96.2 & \textbf{100.0} \\
LatentVLA \citeyear{xie2026flarelearningfutureawarelatent} & 92.4 & 98.9 & 98.2 & 88.2 & 96.0 & \textbf{100.0} \\
\midrule
\multicolumn{7}{l}{\textbf{End-to-End Methods}} \\
DrivingGPT \citeyear{chen:24drivinggpt} 
& 82.4 & 98.9 & 90.7 & 79.7 & 94.9 & 95.6 \\
UniAD \citeyear{hu2023planningorientedautonomousdriving}  & 83.4 & 97.8 & 91.9 & 78.8 & 92.9 & \textbf{100.0} \\
DriveX-S \citeyear{shi:25drivex} & 84.5 & 97.5 & 94.0 & 79.7 & 93.0 & \textbf{100.0}\\
World4Drive \citeyear{zheng:25world4drive} & 85.1 & 97.4 & 94.3 & 79.9 & 92.8 & \textbf{100.0} \\
DRAMA \citeyear{zhang:25drama} & 85.5 & 98.0 & 93.1 & 80.1 & 94.8 & \textbf{100.0} \\
VADv2 \citeyear{jiang2026vadv2endtoendvectorizedautonomous} & 86.2 & 98.1 & 94.8 & 80.6 & 94.3 & \textbf{100.0} \\
PRIX \citeyear{wozniak:26prix} & 87.8 & 98.1 & 96.3 & 82.3 & 94.1 & \textbf{100.0} \\
DiffusionDrive \citeyear{liao2025diffusiondrivetruncateddiffusionmodel} & 88.1 & 98.2 & 96.2 & 82.2 & 94.7 & \textbf{100.0} \\
DIVER \citeyear{xia2026diverdivingdeeperdistilleddata} & 88.3 & 98.5 & 96.5 & 82.6 & 94.9 & \textbf{100.0}\\
TrajDiff \citeyear{gui2025trajdiffendtoendautonomousdriving} & 88.5 & 98.1 & 97.0 & 82.7 & 94.3 & \textbf{100.0} \\
Hydra-MDP++ \citeyear{li2025hydramdpadvancingendtoenddriving} & 91.0 & 98.6 & 98.6 & 85.7 & 95.1 & \textbf{100.0} \\
iPad \citeyear{guo2025ipaditerativeproposalcentricendtoend} & 91.7 & 98.6 & 98.3 & 88.0 & 94.9 & \textbf{100.0} \\
Centaur \citeyear{sima2025centaurrobustendtoendautonomous} & 92.6 & \textbf{99.5} & 98.9 & 85.9 & \textbf{98.0} & \textbf{100.0} \\
DriveSuprim \citeyear{yao2025drivesuprimprecisetrajectoryselection} & 93.5 & 98.6 & 98.6 & 91.3 & 95.5 & \textbf{100.0} \\
DrivoR \citeyear{drivor} & 93.7 & 99.0 & 98.9 & 90.0 & 96.7 & \textbf{100.0} \\
RAP-DINO \citeyear{feng2026rap3drasterizationaugmented} & 93.8 & 99.1 & 98.9 & 90.3 & 96.7 & \textbf{100.0} \\
\rowcolor{gray!20} \textbf{RRDrive (Ours)} & \textbf{95.1} & 99.3 & 99.0 & \textbf{92.2} & 97.6 & \textbf{100.0} \\
\bottomrule

\end{tabular}
\caption{Comparison of results on the NAVSIM benchmark. All metrics are higher-is-better. Best results are highlighted in bold. VLA-based and E2E methods are sorted ascendingly by PDMS. $\uparrow$~/~$\downarrow$: higher~/~lower is
better.}
\label{tab:navsim_compare_sorted}
\end{table}

\subsection{Experimental Setup}
We evaluate on NAVSIM \emph{navtest} ($12{,}146$ scenes, $64$ candidates each),
reporting PDMS and its submetrics (NC, DAC, DDC, TTC, EP, C). Candidates are graded
by the oracle PDMS $q$ as \emph{bad} ($q\!<\!0.3$), \emph{poor}
($0.3\!\le\!q\!<\!0.7$), or \emph{good} ($q\!\ge\!0.7$); all methods take the scorer
argmax without re-ranking, and the baseline is a faithfully reproduced DrivoR under
the same protocol. The external RiskOcc predictor is trained offline by
\begin{equation}
\mathcal{L}=\mathcal{L}_{s}+\tfrac{1}{2}\mathcal{L}_{\ell}+\tfrac{3}{2}\mathcal{L}_{d}
+\tfrac{1}{5}\mathcal{L}_{v}+\tfrac{3}{10}\mathcal{L}_{\mathrm{cam}},
\end{equation}
with static/lane heads using focal-CE, Dice, and smooth-$\ell_1$/SSIM, and a dense
dynamic $\mathcal{L}_{d}$ combining CE, Dice, Lov\'asz, a weighted foreground BCE
for recall, a false-positive-penalizing Tversky
($0.3\,\mathrm{FN}{+}0.7\,\mathrm{FP}$), and an AdaNorm auxiliary.

\subsection{Main Results}
Table~\ref{tab:navsim_compare_sorted} reports PDMS and its submetrics against recent VLA-based and end-to-end planners on NAVSIM \emph{navtest}. Placing our register-based planner at the top of the end-to-end group confirms that closing the candidate-quality and ranking gaps identified in the abstract yields a state-of-the-art final score, with the margin driven by the ego-progress (EP) and time-to-collision (TTC) terms while the safety submetrics stay saturated.

\subsection{Gain on Candidate Generation Quality}
Table~\ref{tab:gen} evaluates the candidate pool---good/bad ratios, mean
candidate PDMS, and good candidates per scene---independently of the scorer, isolating
the abstract's first bottleneck (a low candidate-quality floor) and the risk-conditioned
generation module built to raise it. On these metrics RiskOcc raises the \emph{good ratio} (fraction of
candidates with $q\!\ge\!0.7$) by $8.4$ percentage points, nearly halves the
\emph{bad ratio} ($q\!<\!0.3$), improves the \emph{mean candidate PDMS} from $0.797$
to $0.858$, and increases the number of \emph{good candidates per scene} from $51.4$ to $56.8$. 

\begin{figure}[h!]
\centering
\includegraphics[width=0.48\textwidth,    
        trim=10 25 10 20,
        clip]{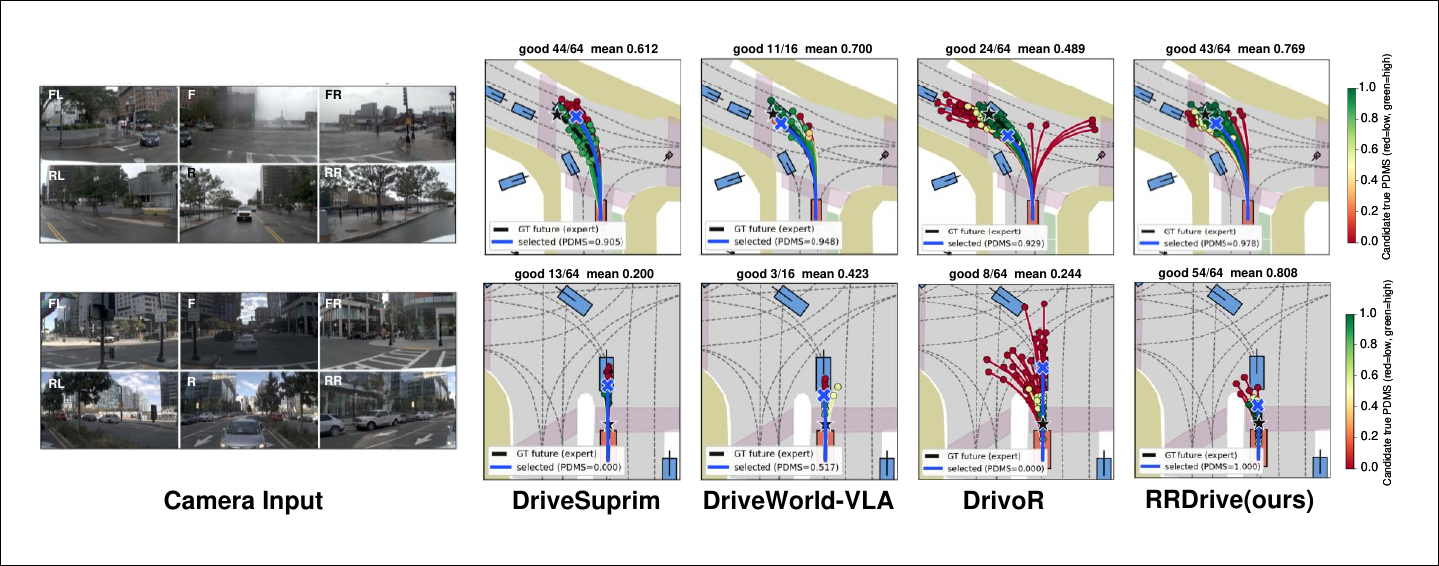} 
\caption{Qualitative comparison of candidate generation on two generation-challenging scenes. Compared with DrivoR (third column), RRDrive (fourth column) concentrates the finite proposal budget within feasible, high-quality regions aligned with the expert trajectory, producing more good candidates and higher mean candidate PDMS.}
\label{fig:generation}
\end{figure}

\begin{table*}[t]
\centering
\small
\setlength{\tabcolsep}{7pt}
\begin{tabular}{@{}ll cccccc c c@{}}
\toprule
\textbf{Variant} & \textbf{Module} & \textbf{NC\,$\uparrow$} & \textbf{DAC\,$\uparrow$} & \textbf{DDC\,$\uparrow$} & \textbf{TTC\,$\uparrow$} & \textbf{EP\,$\uparrow$} & \textbf{Comf.\,$\uparrow$} & \textbf{PDMS\,$\uparrow$} & \textbf{$\Delta$} \\
\midrule
Baseline (no RiskOcc)                & --    & 99.02 & 98.92 & 97.23 & 96.72 & 89.91 & 100.00 & 93.67 & --      \\
\;+ global RiskOcc encoder              & Gen.  & 99.24 & \textbf{99.08} & 97.51 & 97.34 & 90.96 & 99.99  & 94.49 & $+0.82$ \\
\;+ trajectory-conditioned sampling  & Rank. & 99.30 & 99.02 & \textbf{97.56} & 97.57 & 91.25 & 99.99  & 94.68 & $+0.19$ \\
\;+ bidirectional cross-modal fusion & Gen.  & \textbf{99.37} & 99.04 & 97.46 & \textbf{97.78} & 91.51 & 100.00 & 94.90 & $+0.22$ \\
\;+ TTC/EP \& RiskOcc refine          & Rank. & 99.34 & 98.98 & 97.41 & 97.55 & \textbf{92.19} & \textbf{100.00} & \textbf{95.06} & $+0.16$ \\
\bottomrule
\end{tabular}
\caption{Cumulative ablation on navtest (argmax, no re-ranking), decomposed
into all PDMS sub-metrics. \textbf{Module} tags the \emph{primary} pathway each
component serves: Gen.\ = candidate generation, Rank.\ = candidate ranking.
All metrics are higher-is-better; NC = no at-fault collision, DAC = drivable-area
compliance, DDC = driving-direction compliance, TTC = time-to-collision,
EP = ego-progress, Comf.\ = comfort. $\Delta$ is the absolute PDMS gain over the
preceding row. Rows are model-lineage proxies: each step carries minor
co-changes, so tags denote the dominant pathway rather than an exclusive one.}
\label{tab:ablation}
\end{table*}
\label{sec:ablation}

This reflects a \emph{concentration} rather than a diversification of the
candidate budget: among good candidates, the endpoint spread shrinks from $2.81$ to
$1.55$\,m and the number of distinct endpoint modes drops from $20.7$ to $16.2$, i.e.\
RiskOcc reallocates proposals from infeasible behaviors toward the feasible
high-quality region. The qualitative comparison in Figure~\ref{fig:generation} confirms that RRDrive overcomes the low-quality candidate-generation failure by concentrating the finite proposal budget within feasible, high-quality motion regions rather than dispersing it across unsafe or ineffective branches. In the two generation-challenging scenes, the fourth column produces compact candidate bundles aligned with the expert trajectory, yielding $43/64$ and $54/64$ good candidates with mean PDMS values of $0.769$ and $0.808$, whereas the DrivoR candidates in the third column diverge toward multiple low-score branches and contain only $26/64$ and $8/64$ good candidates with mean PDMS values of $0.489$ and $0.244$, respectively.

\begin{table}[h!]
\centering\small\setlength{\tabcolsep}{7pt}

\begin{tabular}{>{\columncolor{gray!20}}lc>{\columncolor{gray!10}}cc}
\toprule
\textbf{Metric} & \textbf{Baseline} & \textbf{Ours} & \textbf{$\Delta(\%)$} \\
\midrule
Good ratio\,$\uparrow$        & 80.3\% & \textbf{88.7\%} & $+10.5$ \\
Bad ratio\,$\downarrow$       & 13.6\% & \textbf{7.6\%}  & $-44.1$ \\
Mean cand.\ PDMS\,$\uparrow$  & 0.797  & \textbf{0.858}  & $+7.7$  \\
Good cand./scene\,$\uparrow$  & 51.4   & \textbf{56.8}   & $+10.5$ \\
\bottomrule
\end{tabular}
\caption{Candidate generation quality. $\uparrow$~/~$\downarrow$: higher~/~lower is
better; $\Delta(\%)$ is the relative change over the baseline.}
\label{tab:gen}
\end{table}

\subsection{Gain on Candidate Ranking}
Table~\ref{tab:rank} assesses how the scorer orders the $64$ candidates within each
scene---scene-wise Spearman between predicted and true PDMS, oracle recall@$K$ (whether
the true optimum lies in the top $K$), oracle rank of the true optimum (lower is better),
oracle gap (PDMS shortfall of the selected trajectory relative to the in-scene optimum),
and the selected PDMS---directly targeting the abstract's second bottleneck, unreliable
fine-grained ranking, which the trajectory-conditioned local risk pathway is built to fix.
RiskOcc improves every metric: the scene-wise Spearman correlation rises from $0.407$
to $0.584$, oracle recall@1 from $5.1\%$ to $15.4\%$ ($3.0\times$), the oracle rank
from $33.9$ to $27.7$, and the selected PDMS from $0.937$ to $0.951$. To probe the
underlying mechanism, Table~\ref{tab:probe} predicts the good-candidate order from
geometry, risk, or both under scene-grouped cross-validation: the risk profile
contributes substantial ranking information beyond geometry ($0.37\!\to\!0.53$),
whereas risk alone is weak. This is consistent with the weak linear correlation
between simple risk statistics and PDMS ($|r|\!<\!0.08$) and the comparable global
scorer-PDMS Pearson of the two models ($0.57$ vs.\ $0.56$): RiskOcc functions as
nonlinear evidence for the scorer rather than a linear surrogate.

\begin{table}[h!]
\centering\small\setlength{\tabcolsep}{6pt}
\begin{tabular}{>{\columncolor{gray!20}}lc>{\columncolor{gray!10}}cc}
\toprule
\textbf{Metric} & \textbf{Baseline} & \textbf{Ours} & \textbf{$\Delta(\%)$} \\
\midrule
Scene Spearman\,$\uparrow$   & 0.407 & \textbf{0.584} & $+43.5$  \\
Oracle recall@1\,$\uparrow$  & 5.1\%  & \textbf{15.4\%} & $+202.0$ \\
Oracle recall@5\,$\uparrow$  & 12.2\% & \textbf{29.3\%} & $+140.2$ \\
Oracle rank\,$\downarrow$    & 33.9  & \textbf{27.7}  & $-18.3$  \\
Oracle gap\,$\downarrow$     & 0.057 & \textbf{0.034} & $-40.4$  \\
Selected PDMS\,$\uparrow$    & 0.937 & \textbf{0.951} & $+1.5$   \\
\bottomrule
\end{tabular}
\caption{Scene-level ranking quality. $\uparrow$~/~$\downarrow$: higher~/~lower is
better; $\Delta(\%)$ is the relative improvement over the baseline.}
\label{tab:rank}
\end{table}

\begin{figure}[h!]
\centering
\includegraphics[width=0.48\textwidth,
        trim=10 25 10 20,
        clip]{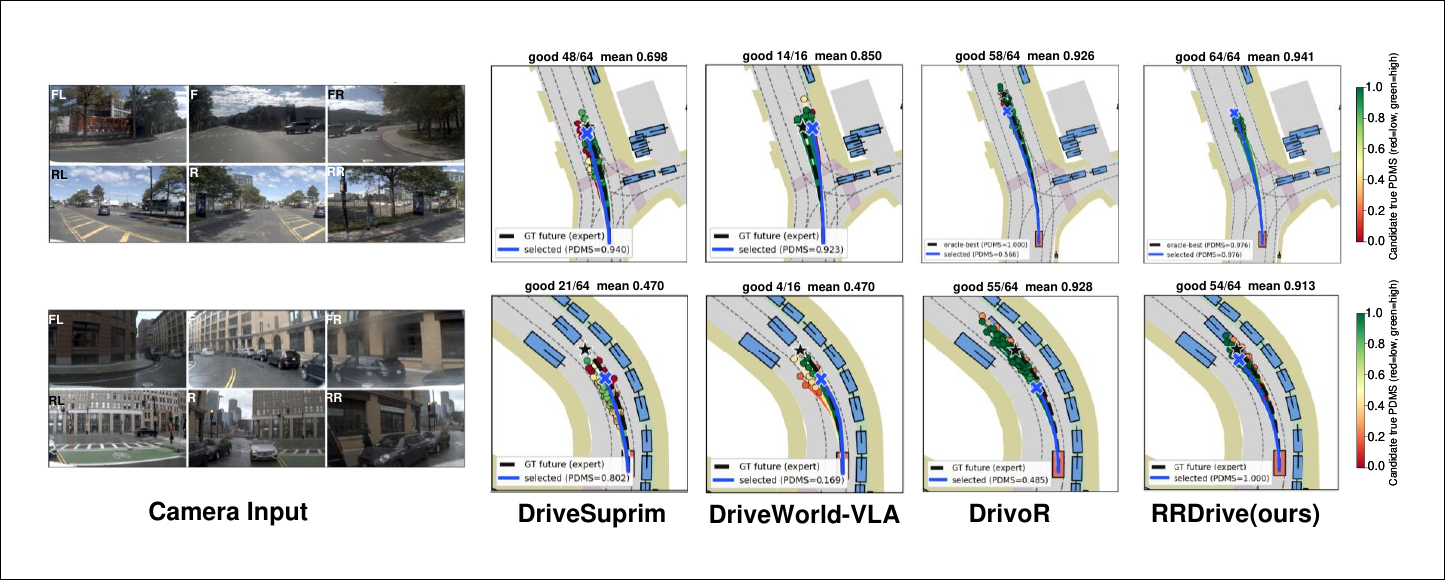} 
\caption{Qualitative comparison of candidate ranking. Within dense bundles of geometrically similar, high-quality candidates, RRDrive selects a near-optimal trajectory where the baseline (third column) mis-ranks and selects a low-scoring one, isolating the improvement to ranking rather than to candidate generation.}
\label{fig:rank}
\end{figure}

\begin{table}[h!]
\centering\small\setlength{\tabcolsep}{5pt}

\begin{tabular}{lcc}
\toprule
\textbf{Feature set} & \textbf{Ego-progress}\,$\uparrow$ & \textbf{True PDMS}\,$\uparrow$ \\
\midrule
Geometry only    & 0.369 & 0.371 \\
RiskOcc only        & 0.174 & 0.177 \\
\rowcolor{gray!20}Geometry + RiskOcc  & \textbf{0.535} & \textbf{0.533} \\
\bottomrule
\end{tabular}
\caption{Within-good-candidate ranking under scene-grouped cross-validation
(mean scene-wise Spearman\,$\uparrow$): the risk profile adds ranking information
beyond trajectory geometry.}
\label{tab:probe}
\end{table}

The qualitative comparison in Figure~\ref{fig:rank} confirms that RRDrive resolves the fine-grained ranking failure by identifying the best trajectory within dense bundles of geometrically similar, high-quality candidates. In the upper scene, the baseline method in the third column already generates $58/64$ good candidates with a mean PDMS of $0.926$ and contains an oracle trajectory with PDMS $1.000$, yet selects a trajectory scoring only $0.566$, while RRDrive selects a $0.976$ trajectory; more decisively, in the lower scene, DrivoR has slightly better pool statistics than RRDrive ($55/64$ good candidates and mean PDMS $0.928$ versus $53/64$ and $0.913$) but selects a $0.485$ trajectory, whereas RRDrive selects the $1.000$ trajectory, isolating the improvement to candidate ranking rather than candidate generation.

\subsection{Difficulty-Stratified Analysis}
Table~\ref{tab:difficulty} partitions the scenes into terciles by baseline difficulty and
reports, for the hardest tercile of each axis, candidate PDMS and good ratio (generation),
scene Spearman and selected PDMS (ranking), and in-good Spearman and oracle recall@1
(confusion), testing the abstract's claim that the gains land where the baseline is
weakest. They do, concentrating in the hardest cases
along both axes. In the hardest \emph{generation} tercile (lowest baseline candidate
PDMS), pool quality improves most; in the hardest \emph{ranking} tercile (largest
baseline oracle gap) and the hardest \emph{confusion} tercile (good candidates most
similar in quality), the ranking metrics improve most, with the within-good Spearman
rising from $0.265$ to $0.674$ and oracle recall@1 from $3.6\%$ to $24.9\%$.

\begin{table}[h!]
\centering\small\setlength{\tabcolsep}{4pt}

\begin{tabular}{l>{\columncolor{gray!20}}lc>{\columncolor{gray!10}}cc}
\toprule
\textbf{Scenario} & \textbf{Metric}\,$\uparrow$ & \textbf{Baseline} & \textbf{Ours} & $\Delta(\%)$ \\
\midrule
Gen-hard & Candidate PDMS   & 0.540  & \textbf{0.703} & $+30.2$  \\
Gen-hard & Good ratio       & 50.1\% & \textbf{69.8\%} & $+39.3$  \\
Rank-hard    & Scene Spearman   & 0.072  & \textbf{0.497} & $+590.3$ \\
Rank-hard    & Selected PDMS    & 0.816  & \textbf{0.884} & $+8.3$   \\
Confu-hard  & In-good Spearman & 0.265  & \textbf{0.674} & $+154.3$ \\
Confu-hard  & Oracle recall@1  & 3.6\%   & \textbf{24.9\%} & $+591.7$ \\
\bottomrule
\end{tabular}
\caption{Hardest-tercile analysis (baseline vs.\ Ours). All listed metrics are
higher-is-better ($\uparrow$); $\Delta(\%)$ is the relative improvement over the
baseline.}
\label{tab:difficulty}
\end{table}

\subsection{Ablation Study}

Table~\ref{tab:ablation} decomposes each RiskOcc component under a strict argmax
protocol (no re-ranking), reporting the full PDMS sub-metric breakdown. Starting from the
reproduced DrivoR baseline (PDMS $0.9367$), we add components cumulatively
and tag each by its primary pathway—generation (Gen.) or ranking (Rank.).

\noindent\textbf{Generation vs.\ ranking contributions.}
The two generation-oriented components—global risk encoding and bidirectional
cross-modal fusion—raise the score by $+0.0104$, while the two ranking-oriented
components—trajectory-conditioned risk sampling and the TTC/EP \& risk-representation
refinement—add a further $+0.0035$, reaching $\mathbf{0.9506}$ for a total gain of
$+0.0139$. Consistent with our two design goals, conditioning generation on the
global RiskOcc field yields the larger share by improving the candidate pool, whereas
the trajectory-aligned risk profile contributes a smaller but complementary gain by
sharpening selection among already-strong candidates.

\noindent\textbf{Where the gain comes from.}
The improvement concentrates in ego-progress (EP $0.8991\!\rightarrow\!0.9219$,
$+0.0228$) and time-to-collision (TTC $0.9672\!\rightarrow\!0.9755$), while the
safety-critical sub-metrics remain saturated throughout (NC $\geq 0.992$,
DAC $\geq 0.989$, DDC $\geq 0.972$, Comfort $=1.000$). The gain therefore stems from
more accurate progress- and TTC-aware generation and selection rather than from
conservative, progress-sacrificing behavior—precisely the fine progress--risk
trade-off that an explicit, trajectory-aligned risk signal is designed to resolve.

\subsection{Robustness to Lossy RiskOcc}
To test whether the framework depends on oracle risk, we train an external
RiskOcc prediction module that infers the risk modality from sensors, yielding a
lossy RiskOcc field in place of the ground truth. As the scorer is trained on clean
risk, we bridge the train/test discrepancy via a lightweight mixed-source
adaptation over ground-truth and predicted RiskOcc. With fully predicted risk,
RRDrive attains a PDMS of $0.9400$, clearly exceeding the no-RiskOcc
baseline ($0.9367$) while trailing the oracle-risk upper bound ($0.9506$). The
residual gap stems mainly from our perception module, whose multi-step temporal
forecasting of RiskOcc remains limited; improving this prediction quality—where
substantial headroom remains—would directly benefit the fusion framework.
These gains under lossy, sensor-derived risk confirm that they arise from the
risk-aware formulation rather than from privileged access to oracle risk.

\section{Conclusion}
We identify compact scene representation as a shared bottleneck of multimodal trajectory generation and selection, and address it with risk-aware occupancy (RiskOcc), a dense, temporally aligned, trajectory-queryable field that unifies static, lane, and dynamic risks. Built on this representation, RRDrive conditions candidate generation on the global risk field and augments the scorer with a trajectory-aligned local risk profile, while RiskOcc-NAVSIM supplies the required supervision through automatic annotation. RRDrive raises candidate quality and ranking reliability, especially in the hardest scenes, and attains a state-of-the-art PDMS of 0.951. An external predictor further verifies that such performance gains persist under sensor-derived risk. The main remaining limitation lies in the multi-step temporal forecasting of dynamic RiskOcc, and improving its prediction quality is a promising direction to fully unlock the representation’s benefits for end-to-end driving.

\begin{small}
\bibliography{aaai2027}
\end{small}


\end{document}


\maketitle

\section{A\quad Rationale of the RiskOcc Value Assignment}
\label{sec:app_value}

\paragraph{Positioning: an ordinal attention prior.}
RiskOcc encodes an \emph{explicit, network-facing risk attention}: a driver-like
measure of how much concern each region of the scene warrants, uniformly quantizing
heterogeneous collision and regulatory risks into a single RiskOcc field that the
planner can read directly. It is thus a subjective attention prior rather than a
physical quantity with a unique ground-truth value, and we neither claim nor require
that the particular numbers are optimal. The only modeling decision is the
\emph{relative ordering} of concern across risk sources; the magnitudes are a
deliberately simple, monotone realization of that ordering.

\paragraph{Priority and observable-cue instantiation.}
The assignment is realized only from \emph{reliably observable} cues: semantic
category, observable height, and velocity. For static occupancy, the per-cell
collision risk is tied to the observable height of the occupying structure in
\emph{positive correlation} (a taller salient structure yields a higher risk value),
rather than to the object's occupied area; for dynamic agents, risk grows with
relative speed and decays over the future horizon. Vulnerable road users
(pedestrians, cyclists) are retained as high-attention dynamic agents rather than
discarded, reflecting the more severe consequences for unprotected humans. We
deliberately avoid latent, unobservable properties such as mass: a heavy versus
light barrier is indistinguishable from sensors, so a mass-conditioned risk would be
unrealizable at both annotation and deployment time. RiskOcc thus trades theoretical
completeness for a representation that is consistent between label construction and
sensor-based prediction.

\paragraph{Identifiability and validation.}
The planner consumes RiskOcc only through trajectory-conditioned sampling and
attention, which respond to the RiskOcc field mainly through its relative ordering
rather than its absolute scale. The assignment is therefore intended to be identified only
\emph{up to a strictly increasing reparameterization} $r\!\mapsto\!\psi(r)$, so
``which exact values'' is not a well-posed question; what must be correct is the
order. The scheme is validated \emph{behaviorally} rather than by hand-tuning:
injecting it improves end-to-end planning, and the gains concentrate exactly where
risk reasoning matters most (the hardest strata; see the cumulative ablation and
difficulty-stratified experiments in the main paper and Appendix~F).
Consistent with this, we observe that replacing the spacing of the value table with
alternative order-preserving schemes leaves the aggregate score essentially
unchanged; exact-value optimality is neither claimed nor needed.

\section{B\quad Implementation Details}
\label{sec:app_impl}

\paragraph{Planner training.}
RRDrive is warm-started from the reproduced DrivoR baseline. Candidate generation
and the trajectory decoder are kept frozen while the risk tokenizer, cross-modal
fusion, and the risk-aware scorer heads are fine-tuned, so the improvement is
attributable to risk conditioning rather than to re-training the generator. The
scorer uses multi-statistic risk pooling (mean/max/min/last over the sampled
footprint profile) and a small learning rate for the calibrated TTC/EP heads;
all reported numbers use the scorer argmax with no re-ranking.

\paragraph{External predictor training.}
The external RiskOcc predictor (Appendix~E) is trained offline with
\begin{equation}
\mathcal{L}=\lambda_{s}\mathcal{L}_{s}+\lambda_{\ell}\mathcal{L}_{\ell}
+\lambda_{d}\mathcal{L}_{d}+\lambda_{v}\mathcal{L}_{v}
+\lambda_{\mathrm{cam}}\mathcal{L}_{\mathrm{cam}},
\end{equation}
where $\lambda_{\bullet}$ balance the static, lane, dynamic, velocity, and
camera-auxiliary terms. The static ($\mathcal{L}_s$) and lane ($\mathcal{L}_\ell$)
heads use focal-CE, Dice, and smooth-$\ell_1$/SSIM, and the dense dynamic term
$\mathcal{L}_d$ combines CE, Dice, Lov\'asz, a weighted foreground BCE for recall,
and a false-positive-penalizing Tversky term.
When the planner is evaluated on predicted RiskOcc, a lightweight mixed-source
adaptation trains it on a stochastic mixture of ground-truth and predicted RiskOcc
(predicted selected with probability $0.7$) to bridge the clean/lossy train--test
gap.

\section{C\quad Metric and Difficulty-Bucket Definitions}
\label{sec:app_defs}

\noindent\textbf{Candidate quality tiers.}\quad
Each candidate is graded by its oracle PDMS $q$: \emph{bad} ($q\!<\!0.3$, in practice
almost always a hard safety/drivability violation with $q\!=\!0$), \emph{poor}
($0.3\!\le\!q\!<\!0.7$), and \emph{good} ($q\!\ge\!0.7$).
\textbf{Ranking metrics.}\quad
For the $64$ candidates of a scene we report \emph{scene Spearman}, the rank
correlation between predicted scores and oracle PDMS; \emph{oracle recall@$K$},
whether the true optimum lies in the scorer's top $K$; \emph{oracle rank}, the
position of the true optimum under the scorer (lower is better); \emph{oracle gap},
the PDMS shortfall of the selected trajectory relative to the in-scene optimum; and
\emph{selected PDMS}, the oracle PDMS of the argmax pick. We additionally report the
\emph{in-good Spearman}, computed \emph{only} over the good candidates
($q\!\ge\!0.7$) of a scene; it measures fine-grained ordering \emph{within} an
already-strong pool and is distinct from the overall scene Spearman.

\noindent\textbf{Formal definitions.}\quad
Let $\mathcal{S}$ be the scene set; each scene has $N\!=\!64$ candidates with oracle
PDMS $q_i\!\in\![0,1]$ and learned score $\hat{s}_i$. The selected and oracle-best
indices are $\hat{\imath}=\arg\max_i \hat{s}_i$ and $i^\star=\arg\max_i q_i$, the
good set is $\mathcal{G}_s=\{i:q_i\!\ge\!\gamma\}$, and $\mathbb{1}[\cdot]$ is the
indicator. With tier thresholds $\tau\!=\!0.3$, $\gamma\!=\!0.7$, the candidate-pool
(generation) metrics (good/valid/bad ratios $\mathrm{GR},\mathrm{VR},\mathrm{BR}$,
mean candidate PDMS $\bar{q}$, and mean good candidates per scene $\bar{n}_g$) are:
\begin{equation}
\begin{aligned}
\mathrm{GR}&=\tfrac{1}{|\mathcal{S}|N}\sum_{s,i}\mathbb{1}[q^s_i\ge\gamma],\\
\mathrm{VR}&=\tfrac{1}{|\mathcal{S}|N}\sum_{s,i}\mathbb{1}[q^s_i\ge\tau],\\
\mathrm{BR}&=\tfrac{1}{|\mathcal{S}|N}\sum_{s,i}\mathbb{1}[q^s_i<\tau],\\
\bar{q}&=\tfrac{1}{|\mathcal{S}|N}\sum_{s,i} q^s_i,\\
\bar{n}_g&=\tfrac{1}{|\mathcal{S}|}\sum_{s,i}\mathbb{1}[q^s_i\ge\gamma].
\end{aligned}
\end{equation}
For ranking, let $d_i=\mathrm{rk}(\hat{s}_i)-\mathrm{rk}(q_i)$ and
$r_s=\bigl|\{j:\hat{s}_j\ge\hat{s}_{i^\star}\}\bigr|$ be the within-scene rank
difference and the scorer rank of the oracle-best. The mean scene Spearman
$\bar{\rho}$, oracle recall@$K$~($\mathrm{R}_{@K}$), mean oracle rank $\bar{r}$,
mean oracle gap $\bar{\delta}$, and mean selected PDMS $\bar{q}_{\hat{\imath}}$ are
\begin{equation}
\begin{aligned}
\rho_s&=1-\tfrac{6\sum_i d_i^2}{N(N^2-1)},\\
\bar{\rho}&=\tfrac{1}{|\mathcal{S}|}\sum_s \rho_s,\\
\mathrm{R}_{@K}&=\tfrac{1}{|\mathcal{S}|}\sum_s\mathbb{1}[r_s\le K],\\
\bar{r}&=\tfrac{1}{|\mathcal{S}|}\sum_s r_s,\\
\bar{\delta}&=\tfrac{1}{|\mathcal{S}|}\sum_s\bigl(q^s_{i^\star}-q^s_{\hat{\imath}}\bigr),\\
\bar{q}_{\hat{\imath}}&=\tfrac{1}{|\mathcal{S}|}\sum_s q^s_{\hat{\imath}}.
\end{aligned}
\end{equation}
The \emph{in-good} Spearman evaluates $\rho_s$ over $\mathcal{G}_s$ (scenes with
$|\mathcal{G}_s|\!\ge\!2$), and the difficulty buckets use the relative oracle gap
$(q^s_{i^\star}\!-\!q^s_{\hat{\imath}})/q^s_{i^\star}$ (set to $0$ when
$q^s_{i^\star}\!=\!0$).
\textbf{Two Spearman calibers.}\quad
We stress that two Spearman numbers appear in this work and are \emph{not}
interchangeable: (i) the \emph{overall} scene Spearman over all $64$ candidates,
averaged over all scenes ($0.407\!\to\!0.584$ in the main paper), and (ii) the
\emph{in-good} Spearman restricted to good candidates within the hardest
(high-confusion) stratum ($0.265\!\to\!0.674$, Appendix~F).
The latter is larger because it isolates the fine-grained regime the local risk
pathway targets.

\noindent\textbf{Difficulty buckets.}\quad
We sort the $12{,}146$ navtest scenes by a baseline-derived difficulty score and
split into terciles (low $30\%$ / mid $40\%$ / hard $30\%$, i.e.\ $3644/4858/3644$
scenes); both models are bucketed by the \emph{same} baseline split for a consistent
comparison. Three axes are used: (a) \emph{generation} difficulty, by ascending
baseline mean candidate PDMS (lowest is hardest); (b) \emph{ranking} difficulty, by
descending baseline relative oracle gap $\mathrm{gap}/\mathrm{oracle\text{-}PDMS}$
(largest is hardest); and (c) \emph{confusion} difficulty, by ascending baseline
candidate-PDMS standard deviation (smallest spread, i.e.\ the most similar candidates, is
hardest).

\begin{table*}[t]
\centering\small\setlength{\tabcolsep}{5pt}
\begin{tabular}{llcccccc}
\toprule
\textbf{Gen.\ difficulty} & \textbf{Model} & \textbf{Cand.\ PDMS}\,$\uparrow$ & \textbf{Valid}\,$\uparrow$ & \textbf{Good}\,$\uparrow$ & \textbf{Spearman}\,$\uparrow$ & \textbf{Rel.\ gap}\,$\downarrow$ & \textbf{Sel.\ PDMS}\,$\uparrow$ \\
\midrule
Low  & Baseline & 0.962 & 98.2\% & 98.1\% & 0.312 & 0.011 & 0.989 \\
Low  & Ours     & 0.959 & 98.8\% & 98.8\% & 0.590 & 0.005 & 0.988 \\
Mid  & Baseline & 0.867 & 89.8\% & 89.4\% & 0.353 & 0.045 & 0.951 \\
Mid  & Ours     & 0.899 & 95.6\% & 95.4\% & 0.617 & 0.018 & 0.966 \\
Hard & Baseline & 0.540 & 50.5\% & 50.1\% & 0.570 & 0.122 & 0.866 \\
Hard & Ours     & \textbf{0.703} & \textbf{70.0\%} & \textbf{69.8\%} & 0.537 & 0.088 & \textbf{0.892} \\
\bottomrule
\end{tabular}
\caption{Generation-difficulty terciles (bucketed by ascending baseline mean
candidate PDMS). Higher is better except relative gap.}
\label{tab:app_gen}
\end{table*}

\begin{table*}[t]
\centering\small\setlength{\tabcolsep}{6pt}
\begin{tabular}{llccccc}
\toprule
\textbf{Rank.\ difficulty} & \textbf{Model} & \textbf{Spearman}\,$\uparrow$ & \textbf{Rel.\ gap}\,$\downarrow$ & \textbf{Oracle rank}\,$\downarrow$ & \textbf{Recall@1}\,$\uparrow$ & \textbf{Sel.\ PDMS}\,$\uparrow$ \\
\midrule
Low  & Baseline & 0.615 & 0.000 & 26.61 & 10.98\% & 0.995 \\
Low  & Ours     & 0.620 & 0.010 & 26.78 & 13.53\% & 0.984 \\
Mid  & Baseline & 0.512 & 0.013 & 30.89 & 4.38\%  & 0.984 \\
Mid  & Ours     & 0.629 & 0.014 & 24.69 & 18.92\% & 0.976 \\
Hard & Baseline & 0.072 & 0.175 & 45.33 & 0.00\%  & 0.816 \\
Hard & Ours     & \textbf{0.497} & \textbf{0.089} & \textbf{32.58} & \textbf{12.57\%} & \textbf{0.884} \\
\bottomrule
\end{tabular}
\caption{Ranking-difficulty terciles (bucketed by descending baseline relative
oracle gap). Higher is better except relative gap and oracle rank.}
\label{tab:app_rank}
\end{table*}

\section{D\quad Risk-Aware PDM Scorer Architecture}
\label{sec:app_scorer}

\begin{figure}[h!]
    \centering
    \includegraphics[width=0.99\linewidth]{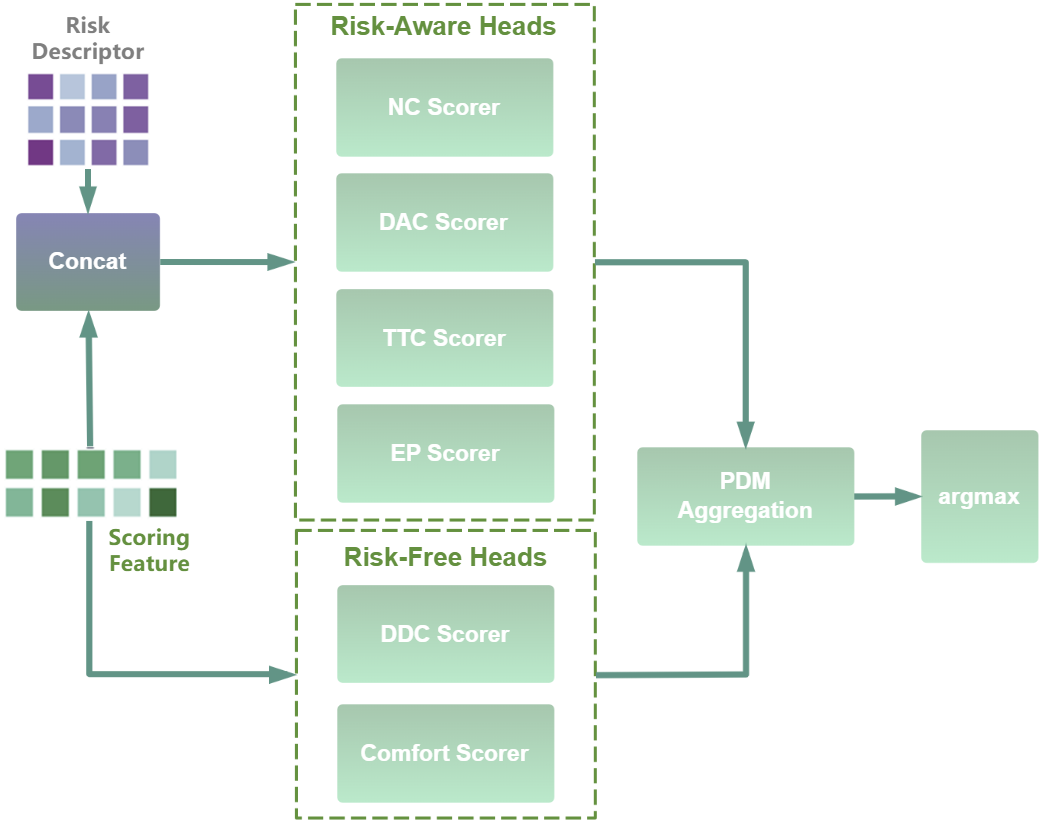}
    \caption{Risk-aware PDM scorer architecture. For each candidate, the
    trajectory-conditioned \emph{risk descriptor} is concatenated with the
    scorer's \emph{scoring feature} and fed to the risk-aware heads
    (NC, DAC, TTC, EP), whereas the risk-free heads (DDC, comfort) read the
    scoring feature alone. All sub-metric scores are combined by PDM weighted-log
    aggregation, and the best candidate is selected by argmax.}
    \label{fig:app_scorer}
\end{figure}

Figure~\ref{fig:app_scorer} details the risk-aware PDM scorer that realizes the
trajectory-conditioned candidate selection of the main paper. For each candidate,
the scorer attention over the shared fused context produces a \emph{scoring
feature} $u_i$, while the ego footprint sampled along the time-aligned RiskOcc
field yields a \emph{risk descriptor} $g_i$---the multi-statistic
(mean/max/min/last) pooling of the per-waypoint exposure profile. The two are
concatenated into $[u_i,g_i]$ and routed only to the \emph{risk-aware heads}
(NC, DAC, TTC, EP), whose scores depend on the trajectory-specific collision,
drivable-area, time-to-collision, and progress risk that the local risk profile
exposes. The remaining \emph{risk-free heads} (DDC and comfort) read the scoring
feature $u_i$ alone, since they are governed by trajectory geometry and dynamics
rather than by scene risk.

All six sub-metric heads feed the standard PDM
weighted-log aggregation, and the final trajectory is chosen by
$\arg\max_i s_i$. This design injects risk evidence exactly where it is
discriminative while leaving the geometry- and comfort-driven sub-metrics
unchanged, so the sharpened ranking among geometrically similar good candidates
is attributable to the risk descriptor rather than to a re-trained scorer
backbone.

\section{E\quad External RiskOcc Predictor: Architecture and Robustness}
\label{sec:app_pred}

\begin{figure}[t]
    \centering
    \includegraphics[width=0.99\linewidth]{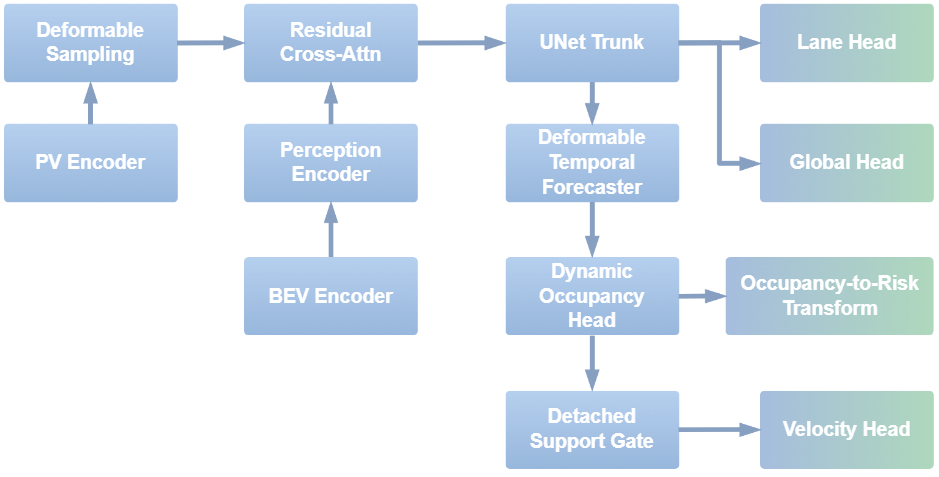}
    \caption{The architecture of the external RiskOcc predictor:
    LiDAR BEV encoder with temporal aggregation, cross-modal fusion by residual
    spatial cross-attention, and a class-first risk-occupancy decoder.}
    \label{fig:app_pred_arch}
\end{figure}

Injecting a temporal, BEV-based occupancy branch \emph{inside} the planner would
break its lightweight, history-free, perspective-view register-token design: DrivoR
maintains neither an explicit temporal history nor a BEV feature space, both of
which occupancy forecasting inherently requires. We therefore keep risk perception
in a \emph{separate} external predictor built on a LiDAR BEV, which natively
provides the metric, ego-centric grid and the short motion history that RiskOcc
prediction needs. This choice matches the paper's scope: our aim is to establish the
\emph{effectiveness and gain} of the RiskOcc representation and the \emph{performance
upper bound} of the fusion framework under clean risk, while the external predictor
supplies a \emph{lossy, sensor-derived} variant solely to probe robustness rather
than to serve as our primary contribution. Given a short sensor observation, the
predictor emits a BEV RiskOcc field through a two-stage encoder--decoder: a LiDAR BEV
encoder with temporal aggregation, a multi-height deformable camera-to-BEV lift,
cross-modal fusion by residual spatial cross-attention into
$\mathbf{B}\in\mathbb{R}^{256\times50\times50}$, and a class-first decoder with
static, lane, and dynamic occupancy-risk heads whose continuous output is
a softmax-class expectation over discrete risk levels.
Figure~\ref{fig:app_pred_arch} details the module internals.

\begin{table*}[t!]
\centering\small\setlength{\tabcolsep}{5pt}
\begin{tabular}{lccccccc}
\toprule
\textbf{Risk source} & \textbf{PDMS}\,$\uparrow$ & \textbf{NC}\,$\uparrow$ & \textbf{DAC}\,$\uparrow$ & \textbf{DDC}\,$\uparrow$ & \textbf{TTC}\,$\uparrow$ & \textbf{EP}\,$\uparrow$ & \textbf{Comf.}\,$\uparrow$ \\
\midrule
None (DrivoR baseline)          & 93.67 & 99.02 & 98.92 & 97.23 & 96.72 & 89.91 & \textbf{100.00} \\
Predicted RiskOcc  & 94.00 & 99.09 & \textbf{99.06} & \textbf{97.53} & 96.89 & 90.30 & \textbf{100.00} \\
GT RiskOcc (upper bound) & \textbf{95.06} & \textbf{99.34} & 98.98 & 97.41 & \textbf{97.55} & \textbf{92.19} & \textbf{100.00} \\
\bottomrule
\end{tabular}
\caption{Robustness to lossy RiskOcc on navtest (argmax, no re-ranking; all metrics
$\times100$, higher is better). A sensor-derived, lossy RiskOcc field still improves
over the no-RiskOcc baseline and recovers roughly a quarter of the
baseline-to-oracle RiskOcc gain.}
\label{tab:app_robust}
\end{table*}


\begin{table*}[t]
\centering
\small
\setlength{\tabcolsep}{5pt}
\begin{tabular}{llcccccc}
\toprule
\textbf{Confusion} & \textbf{Model} & \textbf{Spearman}\,$\uparrow$ & \textbf{In-good Spear.}\,$\uparrow$ & \textbf{Recall@1}\,$\uparrow$ & \textbf{Recall@5}\,$\uparrow$ & \textbf{Oracle rank}\,$\downarrow$ & \textbf{Sel.\ PDMS}\,$\uparrow$ \\
\midrule
Low  & Baseline & 0.598 & -0.014 & 4.01\%  & 13.28\% & 26.04 & 0.912 \\
Low  & \textbf{Ours}     & \textbf{0.538} & \textbf{0.297}  & \textbf{9.82\%}  & \textbf{21.95\%} & \textbf{27.68} & \textbf{0.925} \\
\midrule
Mid  & Baseline & 0.349 & 0.062  & 6.90\%  & 13.28\% & 36.04 & 0.932 \\
Mid  & \textbf{Ours}     & \textbf{0.554} & \textbf{0.510}  & \textbf{12.47\%} & \textbf{25.30\%} & \textbf{31.65} & \textbf{0.954} \\
\midrule
Hard & Baseline & 0.286 & 0.265  & 3.62\%  & 9.58\%  & 39.05 & 0.967 \\
Hard & \textbf{Ours}     & \textbf{0.675} & \textbf{0.674} & \textbf{24.86\%} & \textbf{42.01\%} & \textbf{22.40} & \textbf{0.972} \\
\bottomrule
\end{tabular}
\caption{Confusion terciles (bucketed by ascending baseline candidate-PDMS standard
deviation; the hardest tercile has the most similar candidates). Higher is better
except oracle rank.}
\label{tab:app_confu}
\end{table*}

\begin{figure}[h!]
    \centering
    \includegraphics[width=0.99\linewidth]{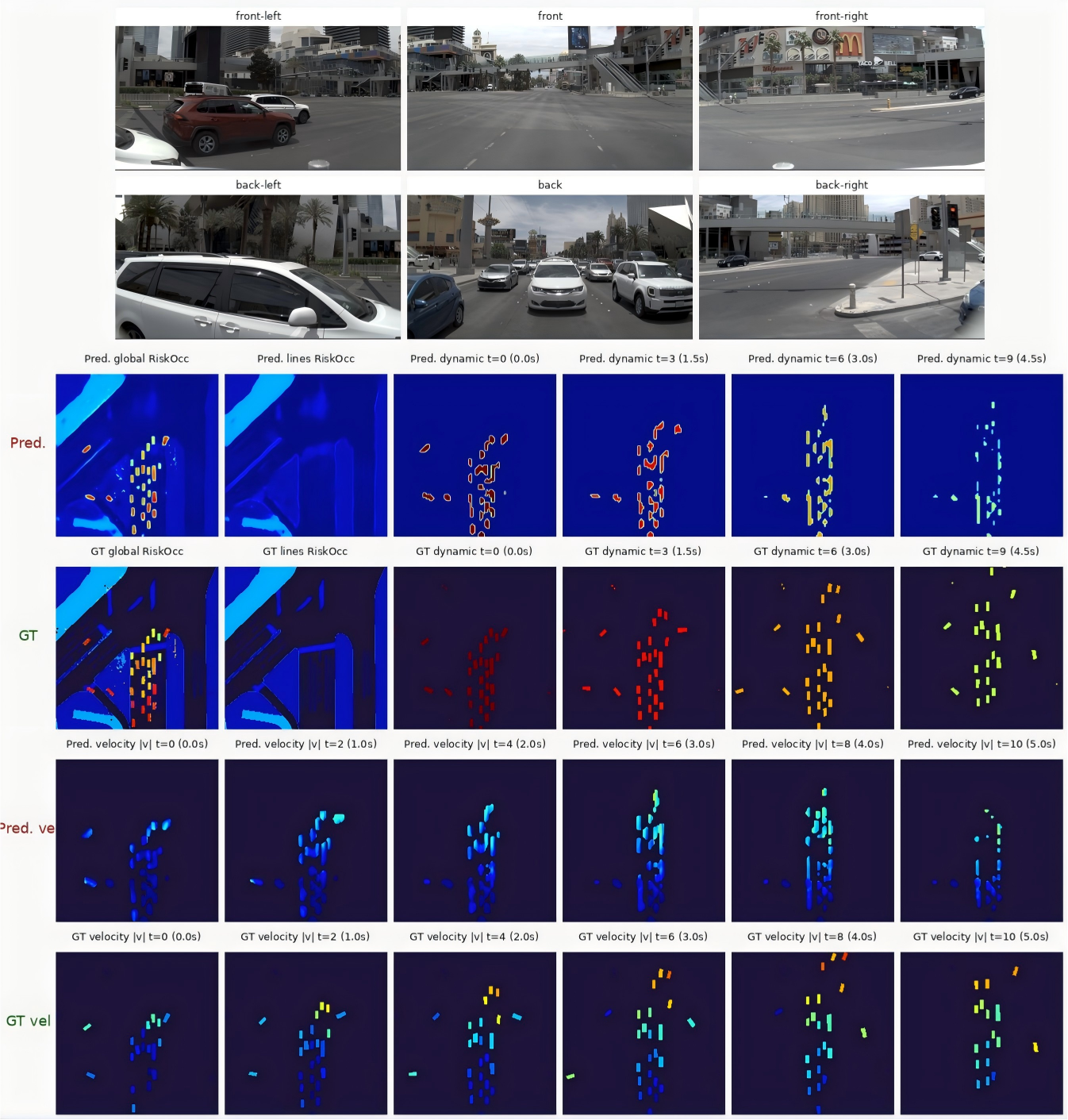}
    \caption{Qualitative predicted RiskOcc from the external predictor (prediction vs. ground truth).
The predicted global and line RiskOcc maps recover the coarse static layout and
lane geometry, although they are smoother than the ground truth. In contrast,
the dynamic RiskOcc is reliable mainly at the current and near-future frames:
at later timesteps, moving-agent footprints become ghosted, blurred, or
partially missing, and the associated velocity maps weaken or become misaligned.
These future-step dynamic errors illustrate a major visible source of the
lossy-input performance drop.}
    \label{fig:app_pred_maps}
\end{figure}

Table~\ref{tab:app_robust} reports the closed-loop robustness result on navtest
(argmax, no re-ranking). With \emph{fully predicted} RiskOcc plus the lightweight
mixed-source adaptation, RRDrive attains PDMS $0.9400$, clearly above the no-RiskOcc
baseline ($0.9367$) while trailing the ground-truth-risk upper bound ($0.9506$);
Figure~\ref{fig:app_pred_maps} localizes this residual gap. The net improvement over
the baseline confirms that the gains arise from the risk-aware formulation rather
than from privileged access to oracle risk.

Figure~\ref{fig:app_pred_maps} isolates \emph{where} the predicted RiskOcc field
loses fidelity. The global RiskOcc is predicted accurately and closely tracks the ground
truth, so the generation-side conditioning is largely preserved under prediction.
The dynamic RiskOcc, however, is reliable only for the near horizon; at later
timesteps the forecast smears moving agents into ghosted, blurred occupancy, the
very signal the selection-side pathway reads along each candidate trajectory. This
mismatch between clean-risk training and blurred predicted dynamics is the principal
reason the fully-predicted variant trails the ground-truth upper bound in
Table~\ref{tab:app_robust}. Sharpening multi-step dynamic RiskOcc forecasting (e.g.\
explicit motion modeling that suppresses temporal diffusion) is therefore a
central direction for future work and would directly narrow the
remaining gap.

\section{F\quad Difficulty-Stratified Full Results}
\label{sec:app_difficulty}

Tables~\ref{tab:app_gen}, \ref{tab:app_rank}, and~\ref{tab:app_confu} give the full
per-tercile breakdown (the main paper reports only the hardest tercile of each axis). Gains are largest in
the hardest terciles: generation quality improves most where the baseline pool is
weakest, while the ranking and in-good ordering improve most where the baseline
scorer is most confused.

\noindent\textbf{Generation terciles (Table~\ref{tab:app_gen}).}
The effect is monotone in difficulty. In the easy tercile the baseline pool is
already near-saturated ($0.962$ mean candidate PDMS, $98.1\%$ good), leaving no
headroom, so our pool statistics are essentially unchanged and only the scene
Spearman rises markedly ($0.312\!\to\!0.590$). Gains grow through the mid tercile
(mean candidate PDMS $0.867\!\to\!0.899$, good ratio $89.4\%\!\to\!95.4\%$, selected
PDMS $0.951\!\to\!0.966$) and peak in the hard tercile, where the baseline generator
collapses ($0.540$ mean, $50.1\%$ good): RRDrive raises the mean candidate PDMS by
$+30.2\%$ to $0.703$, the good ratio to $69.8\%$, and the selected PDMS to $0.892$.
We note honestly that in this hardest tercile the \emph{full-scene} Spearman dips
slightly ($0.570\!\to\!0.537$): when the baseline pool is dominated by bad
candidates, coarse ordering is comparatively easy, so our gain here comes from a
substantially stronger pool and higher final selection rather than from ordering.

\noindent\textbf{Ranking terciles (Table~\ref{tab:app_rank}).}
The shape is opposite: the effect is negligible on easy scenes and decisive on hard
ones. In the easy tercile the baseline scorer already orders candidates well
(Spearman $0.615$, relative gap $\approx\!0$), so there is little to fix and our
selected PDMS is marginally lower ($0.995\!\to\!0.984$) only because the baseline is
near-perfect. The mid tercile already shows clear gains (Spearman
$0.512\!\to\!0.629$, oracle rank $30.9\!\to\!24.7$, recall@1 $4.4\%\!\to\!18.9\%$).
The hard tercile is the most telling: the baseline scorer is essentially random
(Spearman $0.072$, oracle rank $45.3$, recall@1 $0\%$), and RRDrive restores it to a
usable ranker (Spearman $0.497$, oracle rank $32.6$, recall@1 $12.6\%$, selected
PDMS $0.816\!\to\!0.884$), exactly the failure mode the trajectory-conditioned
local-risk pathway is built to repair.

\noindent\textbf{Confusion terciles (Table~\ref{tab:app_confu}).}
This axis most directly probes fine-grained ranking, since the hardest tercile
contains scenes whose $64$ candidates are nearly equal in true quality. The in-good
Spearman (ordering \emph{within} the good pool) improves across all terciles and
most strongly on the hardest one ($0.265\!\to\!0.674$), with the full scene Spearman
rising in step ($0.286\!\to\!0.675$), oracle recall@1 from $3.6\%$ to $24.9\%$,
recall@5 from $9.6\%$ to $42.0\%$, and oracle rank almost halved
($39.1\!\to\!22.4$). The very low baseline in-good Spearman on the easy tercile
($-0.014$) reflects that when candidates are already dissimilar the good subset is
small and its internal order is noisy; RRDrive nonetheless makes this internal order
informative ($0.297$). Together these confirm that the local risk profile sharpens
discrimination precisely where candidates are otherwise indistinguishable.

\section{G\quad Additional Qualitative Results}
\label{sec:app_qual}

\begin{figure*}[thb]
    \centering
    \includegraphics[width=0.99\linewidth]{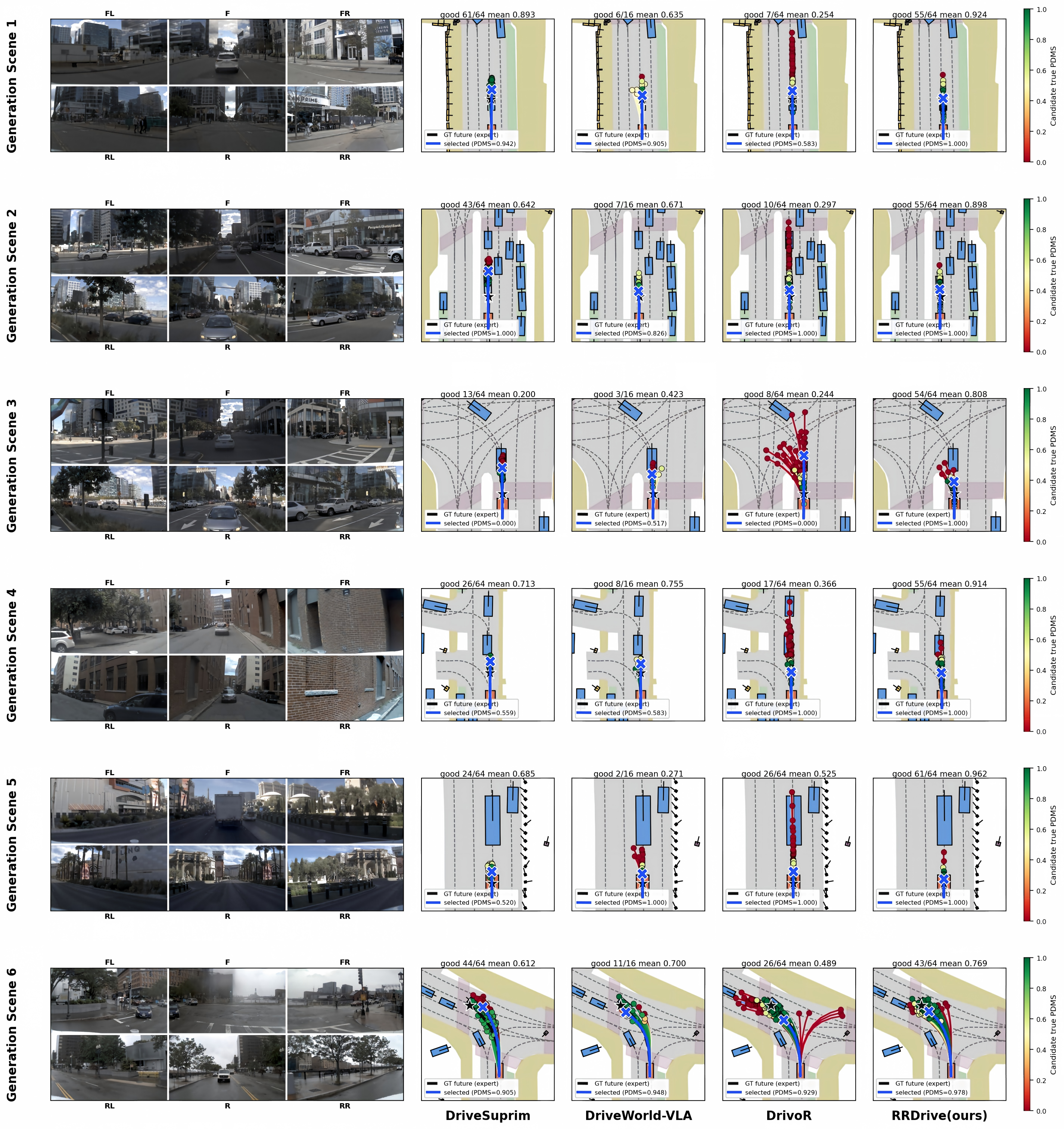}
    \caption{Generation-challenging scenes, comparing DriveSuprim, DriveWorld-VLA,
    the DrivoR baseline, and RRDrive (ours), one method per column. Compared with the
    DrivoR baseline and the other methods, RRDrive concentrates the finite proposal
    budget within feasible, high-quality regions aligned with the expert trajectory,
    yielding more good candidates and higher mean candidate PDMS.}
    \label{fig:appendix1}
\end{figure*}

\begin{figure*}[thb]
    \centering
    \includegraphics[width=0.99\linewidth]{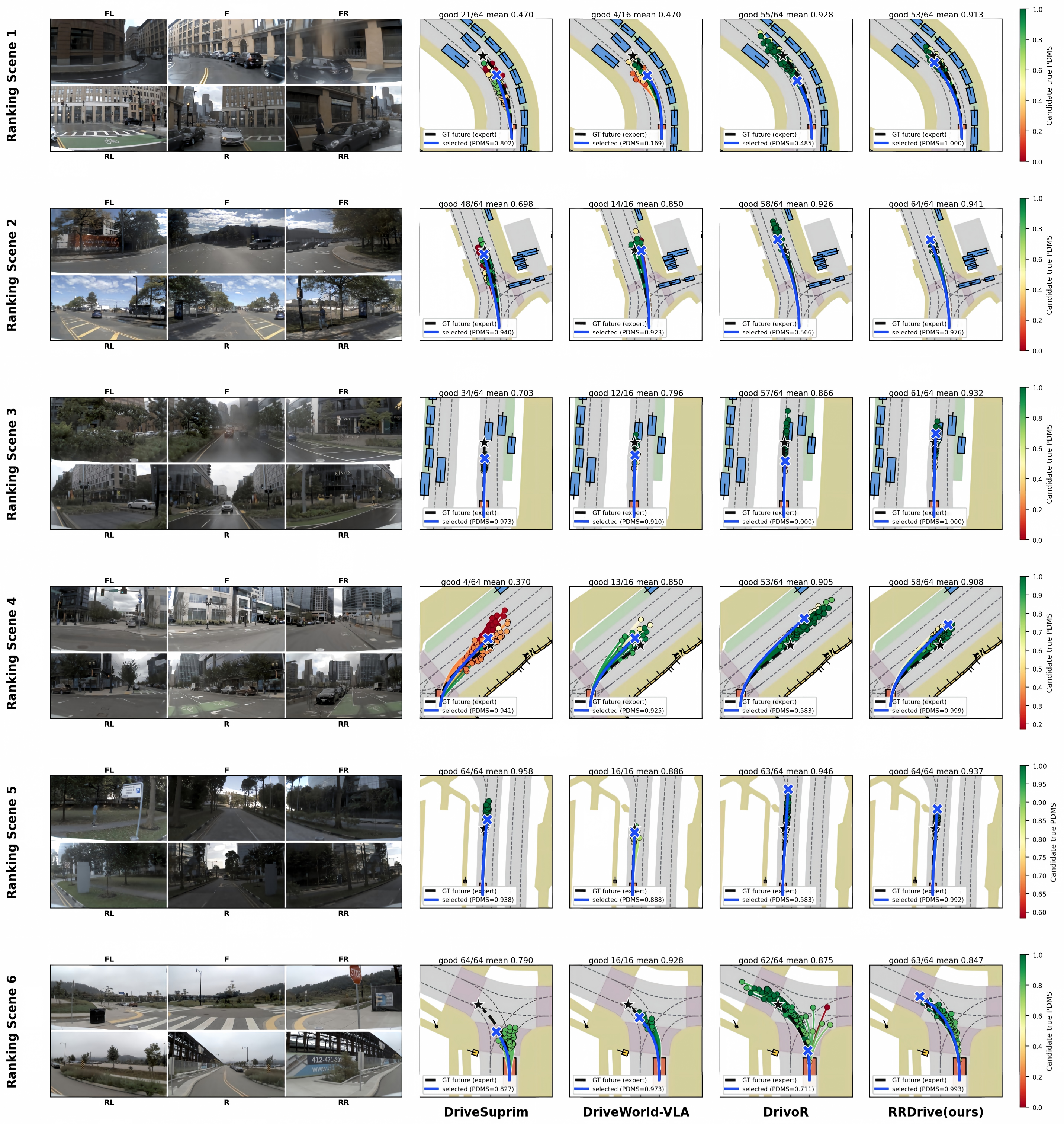}
    \caption{Ranking-challenging scenes, comparing DriveSuprim, DriveWorld-VLA,
    the DrivoR baseline, and RRDrive (ours), one method per column. Within dense
    bundles of geometrically similar, high-quality candidates, RRDrive selects a
    near-optimal trajectory where the DrivoR baseline mis-ranks and selects a
    low-scoring one.}
    \label{fig:appendix2}
\end{figure*}

Figures~\ref{fig:appendix1} and~\ref{fig:appendix2} show additional
generation-challenging and ranking-challenging scenes. Each figure compares four
methods---DriveSuprim, DriveWorld-VLA, the DrivoR baseline, and RRDrive
(ours)---one per column. In each panel the candidate trajectories are overlaid on
the NAVSIM HD map and colored by their oracle PDMS, with the expert (ground-truth)
future and the scorer-selected trajectory highlighted; the surrounding-camera views
are shown on the left.

\noindent\textbf{Candidate generation (Figure~\ref{fig:appendix1}).}
These scenes are generation-challenging: the feasible corridor is narrow, so a
finite proposal budget is easily wasted. The baseline scatters candidates across
unsafe, off-route, or under-progressing directions: a broad fan dominated by
low-PDMS (red) trajectories that only sparsely covers the expert corridor, so that
even a perfect scorer is bounded by a weak pool. Conditioned on the global RiskOcc
field, RRDrive concentrates the same budget inside the feasible, high-quality region
aligned with the expert future, turning most candidates high-PDMS (green) and
raising both the good-candidate fraction and the mean candidate PDMS. The gain is
therefore one of \emph{budget allocation} (covering good behaviors rather than a
low-quality tail), not merely better post-hoc selection.

\noindent\textbf{Candidate ranking (Figure~\ref{fig:appendix2}).}
These scenes are ranking-challenging: both models already produce dense bundles of
geometrically similar, high-quality candidates, so generation is not the
bottleneck. The baseline scorer nonetheless mis-orders these near-identical
trajectories and selects one whose true PDMS is far below the in-scene optimum: its
blue selection departs from both the expert future and the green high-PDMS cluster.
By reading the time-aligned RiskOcc along each candidate's footprint, RRDrive
exposes the fine progress--risk margin that separates them and selects a near-oracle
trajectory that tracks the expert. Because the two candidate pools are comparable
here, the improvement isolates \emph{ranking} rather than candidate quality.

